\documentclass[lettersize,journal]{IEEEtran}
\usepackage{amsmath,amsfonts}
\usepackage{algorithmic}
\usepackage{algorithm}
\usepackage{array}
\usepackage[caption=false,font=normalsize,labelfont=sf,textfont=sf]{subfig}
\usepackage{textcomp}
\usepackage{stfloats}
\usepackage{url}
\usepackage{soul}
\usepackage{verbatim}
\usepackage{graphicx}
\usepackage{multirow}
\usepackage{epstopdf}
\usepackage{cite} 
\usepackage{flushend}
\usepackage{xcolor}
\usepackage{booktabs}
\usepackage{makecell}
\usepackage{hyperref}
\usepackage{cleveref}
\usepackage{placeins}
\usepackage{tabularx}
\usepackage{bm}
\crefname{table}{Table}{Tables}
\Crefname{table}{Table}{Tables}
\crefname{figure}{Fig.}{Figs.}
\Crefname{figure}{Fig.}{Figs.}
\newcommand{\best}[1]{\textbf{\textcolor{red}{#1}}}
\newcommand{\second}[1]{\textbf{\textcolor{blue}{#1}}}
\newcolumntype{C}[1]{>{\centering\arraybackslash}p{#1}}

\usepackage{hyperref}
\hypersetup{
    colorlinks=true,
    linkcolor=red,
    filecolor=blue,      
    urlcolor=blue,
    citecolor=green,
}

\begin{document}

\title{When Sensing Varies with Contexts: Context Probing for 
\\ Tactile Few-Shot Class-Incremental Learning}

\markboth{Journal of \LaTeX\ Class Files,~Vol.~14, No.~8, August~2021}%
{Shell \MakeLowercase{\textit{et al.}}: A Sample Article Using IEEEtran.cls for IEEE Journals}

\author{Yifeng Lin, Aiping Huang, Wenxi Liu, \emph{Senior Member, IEEE}, Si Wu, \emph{Member, IEEE}, Tiesong Zhao, \emph{Senior Member, IEEE}, Zechao Li, \emph{Senior Member, IEEE}, and Zheng-jun Zha, \emph{Member, IEEE}

\thanks{This work is supported by National Natural Science Foundation of China (Grant No. 62571131). Corresponding author: Tiesong Zhao.

Yifeng Lin, Aiping Huang and Tiesong Zhao are with the Fujian Key Laboratory for Intelligent Processing and Wireless Transmission of Media Information, College of Physics and Information Engineering, Fuzhou University, Fuzhou 350108, China. (e-mails: 241120007@fzu.edu.cn, sxxhap@163.com, t.zhao@fzu.edu.cn).

Wenxi Liu is with the College of Computer and Data Science, Fuzhou University, Fuzhou 350108, China. (e-mail: wenxi.liu@hotmail.com).

Si Wu is with the School of Computer Science and Engineering, South China University of Technology, Guangdong, China (e-mail: cswusi@scut.edu.cn).

Zechao Li is with School of Computer Science and Engineering, Nanjing University of Science and Technology, Nanjing 210094, China. (e-mail: zechao.li@njust.edu.cn).

Zheng-jun Zha is with the School of Information Science and Technology, University of Science and Technology of China, Hefei 230027, China. (e-mail: zhazj@ustc.edu.cn).

}}



\maketitle

\begin{abstract}
Few-shot class-incremental learning (FSCIL) aims to recognize novel classes from only a few labeled samples while retaining previously learned knowledge. Although recent FSCIL methods have achieved substantial progress on visual benchmarks, they remain limited in tactile sensing, where the same material may produce markedly different observations under different acquisition contexts, such as sensing devices, contact states, scanning trajectories, and interaction conditions. In tactile FSCIL, the challenges of few-shot learning and class-incremental learning are further amplified by acquisition context: the limited support samples may not only be scarce, but also carry context-induced biases. Once the resulting biased prototypes are inserted into the classifier, they may affect the decision boundaries in subsequent sessions. To address this problem, we propose Context-Probing Few-Shot Class-Incremental Learning (CoP-FSCIL), a context-aware framework for tactile FSCIL. CoP-FSCIL first employs Context-Probing Intervention (CPI) to diagnose local context-sensitive variations in tactile representations. It then introduces a Probe-Conditioned Quotient Adapter (PCQA) to suppress context-sensitive components identified by the probes. Finally, Probe-Stability Prototype Calibration (PSPC) estimates support sample reliability from probe-induced embedding fluctuations and calibrates stochastic prototypes accordingly. Experiments on HapTex and LMT108 show that CoP-FSCIL consistently outperforms representative FSCIL baselines, and extended experiments on audio FSCIL further demonstrate the generality of the proposed context probing mechanism. The source code is currently being prepared and will be released soon.
\end{abstract}


\section{Introduction}

\begin{figure}[!ht]
  \begin{center}
    \centerline{\includegraphics[width=\columnwidth]{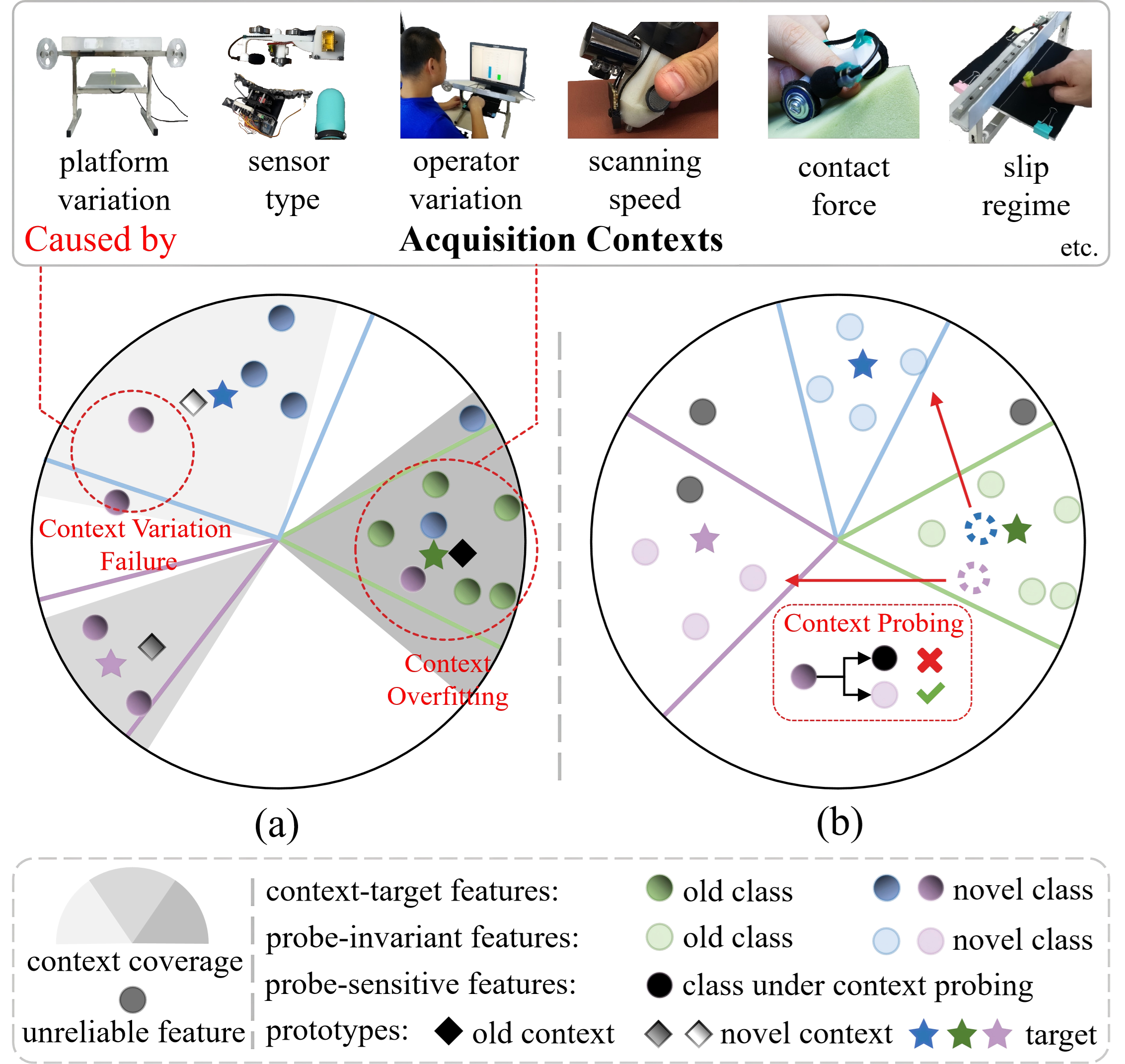}}
    \caption{(a) Acquisition context entangles with target features and degrades FSCIL through two coupled failure modes. Context Overfitting occurs when different classes are covered by similar contexts, causing prototypes to encode context cues and  even yielding unreliable features that are difficult to recover. Context Variation Failure occurs when an unseen context is matched to the closest observed one, resulting in context-dominated prototypes. (b) CoP-FSCIL suppresses context-sensitive components in the embedding space to select context-invariant features and obtain stable representations.
}
\vspace{-1em}
    \label{caotu}
  \end{center}
\end{figure}

\IEEEPARstart{R}{eal-world} intelligent systems rarely operate under a fixed and fully annotated class taxonomy. New categories may appear over time, while collecting sufficient labeled data for each category is often expensive or even impractical. Few-shot class-incremental learning (FSCIL) addresses this open-world learning scenario by requiring a model to learn a set of base classes first and then incorporate novel classes from only a few labeled examples in subsequent incremental sessions. After each session, the model is evaluated over all classes observed so far, which requires both rapid adaptation to novel categories and stable retention of previous knowledge.

Recently, FSCIL has been extended from vision to EEG \cite{li2025anchorinv,cao2025few}, audio \cite{li2025fewaudio}, and other modalities \cite{li2024efficient,wei2025few,xiang2025seeing}. However, these methods share a common assumption: variations among samples of the same class can be implicitly absorbed by the learned embedding. This assumption becomes fragile in tactile sensing. Unlike images captured under relatively standardized visual conditions, tactile observations are strongly coupled with the acquisition process itself. The same material may generate substantially different responses when the sensing device, contact force, scanning speed, contact angle, probe geometry, or interaction trajectory changes. Conversely, different materials acquired under similar contact conditions may exhibit deceptively similar response patterns. Such acquisition-context dependence makes tactile recognition fundamentally different from conventional visual classification and poses a severe challenge to FSCIL. The difficulty is amplified in incremental sessions. Since each novel class is represented by only a few support samples, the estimated prototype may be dominated by context-dependent cues rather than material-intrinsic properties. We refer to this phenomenon as context-induced prototype bias. Once such biased prototypes are added into the classifier, subsequent predictions may rely on context shortcuts, and decision boundaries may become unstable across acquisition conditions. In other words, the main challenge of tactile FSCIL is not only the scarcity of labeled samples, but also the fact that these scarce samples may be context-contaminated. As illustrated in Fig.~\ref{caotu}, this leads to two coupled failure modes: context overfitting, where prototypes encode acquisition-specific cues, and context variation failure, where an unseen context is incorrectly associated with a previously observed context pattern.

Existing FSCIL methods only partially address this issue. Boundary-based methods~\cite{peng2022few,guo2023decision,li2025adaptive} focus on improving classifier updates, but they usually assume that support embeddings are already reliable. Replay-based methods~\cite{agarwal2022semantics,song2023learning,ahmed2024orco,kim2025can} can increase distributional coverage by storing old samples, but their effectiveness depends on memory budgets and does not explicitly distinguish material-intrinsic cues from acquisition-context effects. Prompt-based methods~\cite{park2024pre,liu2025sec,lrt} benefit from large-scale pretraining, yet tactile foundation models and tactile-text supervision remain limited. More importantly, these methods rarely provide a mechanism for identifying which dimensions of a tactile representation are sensitive to acquisition context, especially when explicit context annotations are unavailable.

To bridge this gap, we propose Context-Probing Few-Shot Class-Incremental Learning (CoP-FSCIL), a context-aware framework for tactile FSCIL. The key idea is to actively probe the local context sensitivity of tactile representations through label-preserving interventions. First, Context-Probing Intervention (CPI) synthesizes acquisition-like context probes by combining low-level perturbations and pseudo-context statistic transfer. These probes expose how the embedding changes when the input undergoes plausible context variations. Second, the Probe-Conditioned Quotient Adapter (PCQA) learns a sample-adaptive orthogonal coordinate transformation that separates probe-invariant coordinates from probe-sensitive residuals, allowing the model to suppress context-sensitive components while preserving material-discriminative information. Third, Probe-Stability Prototype Calibration (PSPC) converts probe-induced embedding fluctuations into support-sample reliability weights, thereby reducing the influence of context-unstable samples during prototype construction.

Our contributions are summarized as follows:
\begin{itemize}
\item We propose CoP-FSCIL, a context-probing framework that does not require explicit context labels and uses label-preserving interventions to reveal local context sensitivity in tactile representations.
\item We introduce CPI and PCQA to diagnose and suppress probe-sensitive components through sample-adaptive orthogonal quotient decomposition.
\item We develop PSPC to calibrate stochastic prototypes using probe stability, reducing context-induced prototype bias.
\item Experiments on tactile FSCIL benchmarks demonstrate the superiority of CoP-FSCIL, while extended audio FSCIL results further suggest its generality.
\end{itemize}

\section{Related Work}
\label{Related Work}

\subsection{Few-shot Class-Incremental Learning}

Few-shot class-incremental learning was initially introduced in TOPIC~\cite{tao2020few} and has since increasingly come to adopt a paradigm that decouples feature representation learning from classifier updates~\cite{zhang2021few,zhou2022forward}. Recent studies have moved toward richer model structures and stronger transfer from large-scale pretraining, and can be broadly categorized into boundary-based~\cite{peng2022few,guo2023decision,li2025adaptive}, prompt-based~\cite{park2024pre,liu2025sec,lrt}, and replay-based~\cite{agarwal2022semantics,song2023learning,ahmed2024orco,kim2025can} approaches. Beyond these lines, Comp-FSCIL~\cite{zou2024compositional} explores compositional reasoning by decomposing class knowledge and recombining it for inference, while methods such as Lark~\cite{shi2025lark} and Flexi-FSCIL~\cite{xie2025flexi} focus on continual adaptation via lightweight parameter editing or efficient model fusion. However, as noted above, the majority of FSCIL algorithms are validated primarily on image benchmarks~\cite{krizhevsky2009learning,wah2011caltech,russakovsky2015imagenet}. To address these practical needs, FSCIL has been extended to domains such as physiological signals~\cite{ma2022few,sun2023few,li2025anchorinv}, graphs~\cite{li2024efficient}, and audio~\cite{li2025fewaudio,gao2026tape}. In such multi-modal settings, VT-CIL~\cite{10.1145/3754452} studies class-incremental learning jointly over vision and touch through a unified model with cross-modal regularization. TIFS~\cite{wei2025few} instead leverages tactile cues to synthesize imagined visual observations, using these generated visuals as auxiliary guidance to improve alignment and forgetting resistance. Despite these recent advances, FSCIL that explicitly accounts for acquisition-context effects, particularly in purely tactile sensing, remains largely underexplored.

\subsection{Tactile Datasets}
Tactile datasets vary substantially in sensor type, signal format, interaction protocol, and annotation granularity. Existing datasets can be broadly divided into visual--tactile datasets and numerical--tactile datasets. Visual--tactile datasets, such as VisGel~\cite{visgel}, Touch-and-Go~\cite{touchandgo}, AnyTouch~\cite{anytouch}, and other recent touch-centered multimodal datasets~\cite{tvl,touch100k}, typically record contact-induced deformation images or videos together with visual observations. These datasets are useful for studying geometric representation learning and cross-modal learning through alignment with visual features. However, their observations are closely tied to optical sensor structures, elastomer materials, camera configurations, and contact geometry, and such systems are often more costly and less robust in long-term use. In contrast, numerical--tactile datasets, such as LMT108~\cite{lmt108}, HapTex~\cite{haptex} and Cluster Haptic~\cite{eguchi2026cluster}, mainly collect tactile signals using inertial, force, torque, or related sensors. They usually record one-dimensional or low dimensional time series signals, including acceleration, force, friction, displacement, pressure, audio, and reflectance. Since many of these sensors are built with rigid materials, they are typically cheaper and more durable. Nevertheless, they often rely on scanning platforms that require precise control. For the same material, variations in normal force, scanning speed, motion direction, or operator habit can change the measured tactile response. The raw data of HapTex~\cite{haptex} and LMT108~\cite{lmt108} used in this work are numerical--tactile and are consistent with the IEEE 1918.1.1 standard~\cite{8605315} in practice.

\subsection{Tactile-Based Object Recognition}
Tactile-based recognition has been investigated for material classification, texture recognition, object recognition, and physical-property estimation. Both early and recent studies show that tactile sensing provides complementary information to vision, especially for properties that are difficult to infer from appearance alone, such as roughness, friction, hardness, compliance, and contact state~\cite{liu2017recent,huang2021texture}. With the development of tactile sensors and learning algorithms, convolutional networks have been applied to tactile arrays or pressure images~\cite{gandarias2019cnn}, recurrent models have been used to aggregate temporal contact information~\cite{donato2025tactile}, and multimodal tactile systems have been designed to jointly infer material, curvature, pressure, or other object properties~\cite{xie2024deep,zhao2024augmented}. Beyond closed-set recognition, tactile zero-shot learning further exploits visual or semantic priors to recognize unseen materials or objects~\cite{cao2024multimodal,ueda2024visuo}. Nevertheless, most tactile recognition studies assume a fixed category set and sufficient training data for each class. Their main objective is to improve recognition accuracy under predefined training and testing protocols, rather than to continually incorporate novel tactile categories from only a few samples. In contrast, this work introduces FSCIL into numerical tactile recognition, providing a more realistic setting for tactile systems that must adapt to evolving categories with limited annotation under real deployment constraints.

\section{Method}
\label{Proposed Method}

\subsection{Problem Formulation}
We consider a standard few-shot class-incremental learning protocol consisting of a sequence of sessions ${\ell_0,\ell_1,\ldots,\ell_S}$, where $\ell_0$ denotes the base session and $\ell_i$ ($0 < i\le S$) denotes the $i$-th incremental session. Each session $\ell$ is associated with a label set $\mathcal{Y}_{\ell}$, and the label sets introduced in different sessions are mutually disjoint, i.e., $\mathcal{Y}_{\ell}\cap\mathcal{Y}_{\ell'}=\emptyset$ for $\ell\neq\ell'$. During session $\ell$, the learner is provided with a labeled dataset $\mathcal{D}_{\ell}=\{(x_{i,\ell},y_{i,\ell})\}_{i=1}^{n_{\ell}}$, where $n_{\ell}$ denotes the number of labeled samples, $x_{i,\ell}$ denotes a tactile observation, and $y_{i,\ell}\in\mathcal{Y}_{\ell}$ is the corresponding class label. The base session $\ell_0$ contains relatively sufficient labeled samples and is used to establish the initial representation and classifier over $\mathcal{Y}_{0}$. For each incremental session $\ell\ge 1$, the learner follows an $N$-way $K$-shot protocol. Specifically, $|\mathcal{Y}_{\ell}|=N$, and each novel class $y\in\mathcal{Y}_{\ell}$ is represented by a support set
$S_y=\{(x_{i,\ell},y_{i,\ell})\in\mathcal{D}_{\ell}:\, y_{i,\ell}=y\}$ with $|S_y|=K$. 

The model is updated using the limited data available in the current session and is then evaluated on the cumulative label space $\mathcal{Y}_{\le \ell}=\bigcup_{j=0}^{\ell}\mathcal{Y}_{j}$. Therefore, the evaluation after session $\ell$ simultaneously reflects the ability to recognize newly introduced classes and to preserve discrimination among previously learned classes. In the tactile FSCIL setting studied in this work, each observation $x_{i,\ell}$ is a short tactile signal segment acquired under a specific sensing and interaction condition. Although the class label indicates the material category, the observed signal may also contain acquisition-dependent variations. This distinction is important in few-shot incremental sessions, where each novel class prototype must be estimated from only $K$ support samples.

\begin{figure*}[!ht]
  \centering
  \includegraphics[width=\textwidth]{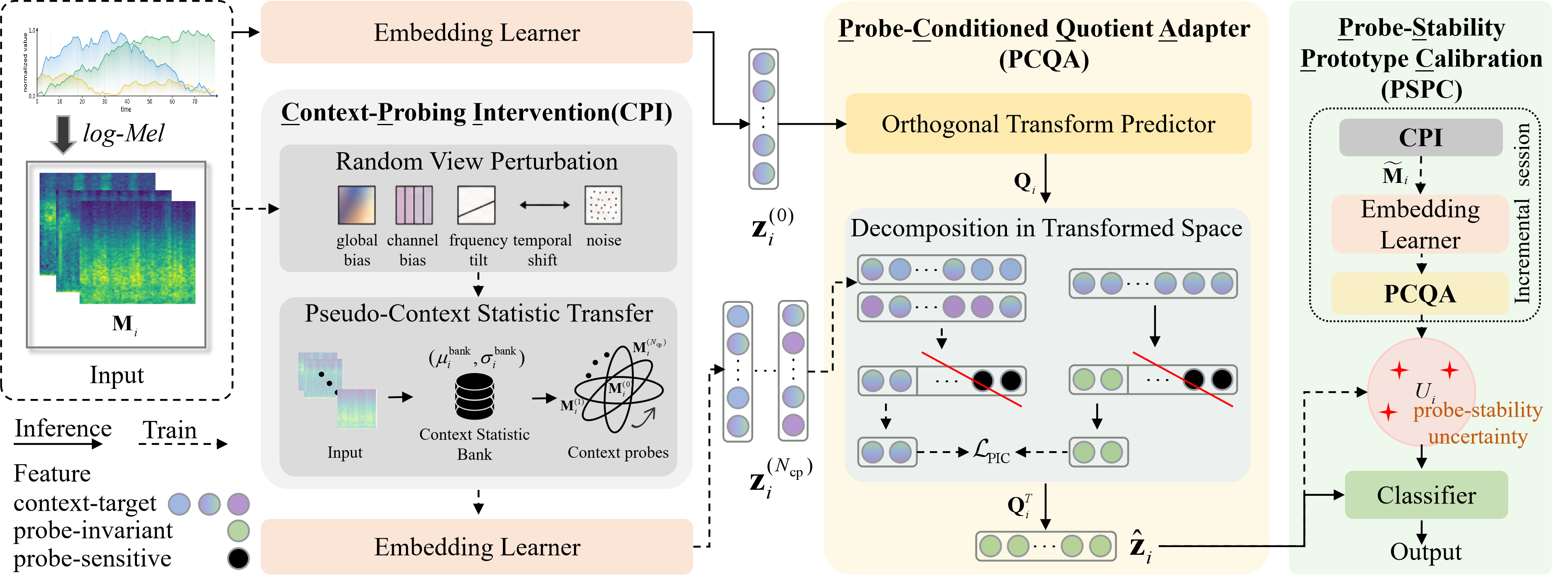}
 \caption{Overall framework of CoP-FSCIL. CoP-FSCIL converts a tactile signal into a log-Mel spectrogram and generates label-preserving context probes through Context-Probing Intervention (CPI). The probe responses are used by the Probe-Conditioned Quotient Adapter (PCQA) to separate probe-invariant and probe-sensitive components, suppressing the latter to obtain a context-robust representation. During incremental learning, Probe-Stability Prototype Calibration (PSPC) estimates support-sample reliability from probe-induced fluctuations and calibrates stochastic prototypes for classification over all observed classes.}
  \label{caotu2}
\end{figure*}

\subsection{Motivation}
Given an input tactile sample $\mathbf{x}$, represented as a short temporal segment of time-series signals, we first convert it into a multi-channel log-Mel spectrogram:
\begin{equation} 
  \mathbf{M} = \mathrm{Mel}(\mathbf{x}),
\end{equation}
For the same material, different acquisition contexts can produce markedly different log-Mel spectrograms. Given an embedding learner \(f_\theta(\cdot)\), the feature representation of a tactile observation \(\mathbf{M}_{y,c}\) from material class \(y\) under acquisition context \(c\) is obtained by applying the learner to the spectrogram:
\begin{equation} 
\mathbf{z}_{y,c} = f_\theta(\mathbf{M}_{y,c}).
\end{equation}
To characterize the influence of acquisition context in the embedding space, we approximate the embedding \(\mathbf{z}_{y,c}\) as:
\begin{equation} 
\mathbf{z}_{y,c} = \bm{\mu}_y^* + \bm{\delta}_c + \bm{\epsilon},
\end{equation}
where \(\bm{\mu}_y^*\) denotes the ideal material-intrinsic center that is independent of acquisition context, \(\bm{\delta}_c\) represents the systematic embedding shift induced by the acquisition context, and \(\bm{\epsilon}\) denotes non-systematic variations and modeling residuals. It is worth noting that this decomposition does not assume that the underlying physical sensing process is linearly additive. Instead, it provides an embedding-space abstraction: acquisition context can displace the material-intrinsic center from its intrinsic position, thereby producing a context-biased prototype. For clarity, we define the context-to-class dispersion ratio $R(y)$ to measure the scale between context-induced intra-class dispersion and intrinsic inter-class center separation:
\begin{equation}
 R(y) = \frac{\max_{c \neq c'} \| \bm{\delta}_c - \bm{\delta}_{c'}\|_2}{ \min_{y' \neq y} \| \bm{\mu}_y^* - \bm{\mu}_{y'}^* \|_2} .
\end{equation}
$R(y)$ measures the largest context-induced embedding dispersion within the same material class relative to the nearest intrinsic center distance to other material classes. As $R(y)$ increases, the context-induced intra-class dispersion gradually approaches, and may even exceed, the inter-class center separation. When the magnitude of intra-class context-induced dispersion becomes comparable to, or even larger than, the separation between inter-class centers, samples from the same material acquired under different contexts may become more dispersed than samples from different materials acquired under similar contexts. This phenomenon is particularly detrimental in the incremental stage, where only a few support samples are available to estimate novel class prototypes, especially when the model encounters previously unseen acquisition contexts. Therefore, it is necessary to suppress the context-sensitive components associated with $\bm{\delta}_c$ during representation learning and prototype construction, so that the model can better preserve material-intrinsic discriminative information.

\subsection{Overall Framework}

We propose CoP-FSCIL, a context-probing framework for Few-Shot Class-Incremental Learning under acquisition-context effects, as illustrated in \cref{caotu2}. Given an observed spectrogram \(\mathbf{M}_{y,c}\), an embedding learner \(f_\theta(\cdot)\), implemented with a standard ResNet-18 backbone, maps it to an embedding \(\mathbf{z}\). Meanwhile, CoP-FSCIL constructs controllable context probes at the input level before feature extraction by combining random view perturbations with pseudo-context statistic transfer. These label-preserving probes reveal how the embedding responds to simulated acquisition-context variations.

Based on the probe responses, PCQA identifies context-sensitive directions in the embedding space and separates them from context-invariant semantic components. Specifically, a lightweight two-layer linear network predicts a set of Householder reflections, which are composed into a sample-adaptive orthogonal transformation. This transformation defines a local coordinate system where probe-invariant and probe-sensitive components are separated. The context-sensitive coordinates are suppressed, and the remaining invariant representation \(\hat{\mathbf{z}}\) is mapped back to the original embedding space through the inverse transformation. The resulting representation is then passed to a classifier \(h_{\theta}\), which predicts labels over all classes observed up to session \(\ell\). For classification, PSPC maintains a stochastic prototype distribution for each class. The classifier samples a prototype \(\tilde{\boldsymbol{\mu}}_{y}\) from
\(\mathcal{N}(\boldsymbol{\mu}_{y}^{\mathrm{pspc}}, \boldsymbol{\sigma}_{y}^{2})\)
and computes cosine-similarity-based class probabilities:
\begin{equation} 
 \begin{aligned}
\tilde{y} = \arg\max_y \frac{e^{\cos(\hat{\mathbf{z}},\tilde{\bm\mu}_y)}}{\sum_{h \in \mathcal{Y}_{\le \ell}} e^{\cos(\hat{\mathbf{z}},\tilde{\bm\mu}_h)}}.
\end{aligned} 
\end{equation}
Unlike fixed prototype classifiers, PSPC maintains learnable per-class centers and scales,
\(\{ \boldsymbol{\mu}_{y}^{\mathrm{pspc}}, \boldsymbol{\sigma}_{y} \}_{y \in \mathcal{Y}_{\le \ell}}\), to model prototype uncertainty throughout incremental learning. CPI-generated probes further estimate the stability of each support sample under controllable context interventions. PSPC uses this stability to down-weight context-sensitive support samples during prototype estimation, thereby reducing prototype bias from context-contaminated embeddings. This is especially important in later incremental sessions, where novel class prototypes are estimated from only a few labeled samples.

\subsection{Context-Probing Intervention}

Context-Probing Intervention is designed to generate label-preserving, context-like probes that reveal how tactile representations respond to acquisition-context variations. Specifically, we combine random view perturbation \(\mathcal{V}\) with pseudo-context statistic transfer \(\mathcal{M}\). Given an input spectrogram \(\mathbf{M}_i\), the context-probing operator \(T_{\mathrm{cp}}(\cdot)\) first applies \(\mathcal{V}\) and then \(\mathcal{M}\), yielding the transformed context probe \(\tilde{\mathbf{M}}_i\):
\begin{equation}
    \tilde{\mathbf{M}}_i
    =
    T_{\mathrm{cp}}(\mathbf{M}_i)
    =
    \mathcal{M}\bigl(\mathcal{V}(\mathbf{M}_i)\bigr).
\end{equation}

\textbf{Random view perturbation} simulates low-level variations caused by changes in the acquisition context. It composes five elementary perturbations: global bias \(b_g\), channel-wise bias \(\mathbf{b}_{\mathrm{ch}}\), frequency-dependent tilt controlled by \(s\), temporal shift \(\Delta t\), and additive noise \(\bm\eta\). Given an input log-Mel spectrogram \(\mathbf{M}_i\), the perturbed view is formulated as follows
\begin{equation}
    \mathcal{V}(\mathbf{M}_i)
    =
    \mathrm{Roll}_{\Delta t}
    \left(
    \mathbf{M}_i + b_g\mathbf{1} + \mathbf{b}_{\mathrm{ch}} + s \mathbf{r}_f + \bm{\eta}
    \right),
\end{equation}
where \(\mathrm{Roll}_{\Delta t}(\cdot)\) denotes a padded temporal shift along the time dimension, and \(\mathbf{r}_f\) is obtained by broadcasting the normalized frequency coordinate \(r_f\) which is defined for frequency index \(f \in \{0,1,\ldots,F-1\}\) over \(F\) Mel bins as
\begin{equation}
    r_f = \frac{2f}{F - 1} - 1 \in [-1,1].
\end{equation}
The global bias models uniform changes in signal energy, while the channel-wise bias captures channel-specific coupling, gain, and contact-strength variations. The frequency tilt introduces a linear change in the spectral envelope along the frequency axis. The temporal shift models scanning-phase variations, and the additive Gaussian noise captures stochastic acquisition disturbances. Unless otherwise specified, \(b_g\), \(\mathbf{b}_{\mathrm{ch}}\), and \(s\) are sampled from uniform distributions, \(\Delta t\) from a discrete uniform distribution, and \(\bm\eta\) from a Gaussian distribution.

\textbf{Pseudo-context statistic transfer} aims to emulate enlarged intra-class context dispersion during context probe construction. For each spectrogram \(\mathbf{M}_i\), we extract its sample-level mean \(\mu_i\) and standard deviation \(\sigma_i\) as compact context-style statistics, computed over the spectrogram plane and channel-wise for multi-channel inputs. These statistics are stored in the Context Statistic Bank as candidate context styles \((\mu_i^{\mathrm{bank}},\sigma_i^{\mathrm{bank}})\), where the bank retains diverse statistics to maintain a dispersed set of context styles during training. Inspired by MixStyle, we normalize the current sample as
\begin{equation}
    \hat{\mathbf{M}}_i =
    \frac{1}{\sigma_i}
    \left(\mathbf{M}_i-\mu_i\mathbf{1}\right).
\end{equation}
We then sample a mixing coefficient \(\lambda_i\) from a Beta distribution and select a reference statistical style
\((\mu_i^{\mathrm{ref}},\sigma_i^{\mathrm{ref}})\) from the banked candidates
\((\mu_i^{\mathrm{bank}},\sigma_i^{\mathrm{bank}})\). The mixed statistics \((\mu_i^{\mathrm{mix}},\sigma_i^{\mathrm{mix}})\) and the resulting pseudo context transferred spectrogram $\tilde{\mathbf{M}}_i$ are computed as
\begin{equation}
\begin{aligned}
     \tilde{\mathbf{M}}_i
    &=
    {\sigma_i^{\mathrm{mix}}\hat{\mathbf{M}}_i
    }
    +
    \mu_i^{\mathrm{mix}}\mathbf{1} ,\\
    \mu_i^{\mathrm{mix}}
    &=
    \lambda_i \mu_i
    +
    (1-\lambda_i)\mu_i^{\mathrm{ref}},\\
    \sigma_i^{\mathrm{mix}}
    &=
    \lambda_i \sigma_i
    +
    (1-\lambda_i)\sigma_i^{\mathrm{ref}}.
\end{aligned}
\end{equation}
Importantly, this transformation does not change the original sample label, and therefore increases intra-class context dispersion while preserving class semantics. To further enlarge intra-class context dispersion, we select the reference style according to the context distance \(d_{ij}\):
\begin{equation}
    d_{ij}
    =
    \left\|
    [\mu_i;\sigma_i]
    -
    [\mu_j^{\mathrm{bank}};\sigma_j^{\mathrm{bank}}]
    \right\|_2,
\end{equation}
where \([\,\cdot\,;\,\cdot\,]\) denotes vector concatenation of the statistical descriptors. We then choose the banked statistics with the largest \(d_{ij}\) as the reference style
\((\mu_i^{\mathrm{ref}},\sigma_i^{\mathrm{ref}})\). If multiple reference styles are required, the candidates are ranked by \(d_{ij}\) and selected accordingly.

It is worth noting that context probes preserve the sample labels. As a result, the model is encouraged to maintain semantic consistency within the context probing, while exposing representation variations caused by acquisition-like changes. In this sense, context probes act as controllable interventions that reveal the local context-sensitive directions of tactile representations. Instead of relying on explicit context labels, we synthesize acquisition-like perturbations and analyze how the resulting embeddings respond in the representation space.

\subsection{Probe-Conditioned Quotient Adapter}

Context-Probing Intervention not only strengthens the robustness of the embedding learner, but also provides explicit probe responses that characterize how the model reacts to acquisition-like context variations. These context-probe responses enable us to examine whether a tactile representation is context-sensitive, and further identify the magnitude and direction of such sensitivity. Based on this observation, the Probe-Conditioned Quotient Adapter formulates context sensitivity analysis as a local coordinate decomposition problem. It learns a set of sample-adaptive coordinate transformations in the representation space, with the goal of separating probe-sensitive components from probe-invariant components and retaining the latter for context-robust classification.

\textbf{Sample-Adaptive Coordinate Observation.}
The proposed module does not aim to remove all probe-induced variations. Instead, it learns a local coordinate system in which probe-invariant coordinates are separated from probe-sensitive residual coordinates. For a given sample \(\mathbf{M}_i\), we generate a set of \(N_\text{cp}\) context probes \(\{\mathbf{M}_i^{(n)}\}_{n=1}^{N_\text{cp}}\), which are encoded into probe embeddings \(\{\mathbf{z}_i^{(n)}\}_{n=1}^{N_\text{cp}}\). The original sample is encoded as \(\mathbf{z}_i^{(0)}\). Conditioned on the original embedding \(\mathbf{z}_i^{(0)}\), we predict a set of \(m\) sample-adaptive reflection column vectors:
\begin{equation}
    \{\mathbf{v}_{i,m}, \mathbf{v}_{i,m-1}, \ldots, \mathbf{v}_{i,1}\}
    =
    g(\mathbf{z}_i^{(0)}),
\end{equation}
where \(g(\cdot)\) is a lightweight context observation network implemented by a two-layer linear network.

We then use these reflection vectors to construct \(m\) Householder reflections. For the \(j\)-th reflection column vector \(\mathbf{v}_{i,j}\), the corresponding Householder matrix is defined as
\begin{equation}
    H(\mathbf{v}_{i,j})
    =
    I
    -
    2
    \frac{
    \mathbf{v}_{i,j} \mathbf{v}_{i,j}^{\top}
    }{
    \mathbf{v}_{i,j}^{\top} \mathbf{v}_{i,j}
    }.
\end{equation}
The sample-adaptive orthogonal transformation is obtained by composing these reflections:
\begin{equation}
    \mathbf{Q}_i
    =
    H(\mathbf{v}_{i,m})
    \cdots
    H(\mathbf{v}_{i,2})
    H(\mathbf{v}_{i,1}).
\end{equation}
Since each Householder reflection is orthogonal, their composition \(\mathbf{Q}_i\) also defines an orthogonal transformation. This transformation provides a sample-specific coordinate system for observing probe induced variations in the embedding space. Since \(\mathbf{Q}_i\) is an orthogonal transformation, it preserves the norm and the underlying inner-product geometry of the embedding space. Therefore, the coordinate transformation itself does not disrupt the stochastic classifier based on cosine similarity. We then map all probe embeddings \(\{\mathbf{z}_i^{(n)}\}_{n=0}^{N_\text{cp}}\) into the sample adaptive coordinate system and the transformed representation is further decomposed into two coordinate blocks:
\begin{equation}
\begin{aligned}
    \mathbf{p}_i^{(n)}
    =
    \mathbf{Q}_i \mathbf{z}_i^{(n)}&,\\
    \left[
    \mathbf{p}_{i,\mathrm{inv}}^{(n)};
    \mathbf{p}_{i,\mathrm{sen}}^{(n)}
    \right]
    =
    &\mathbf{p}_i^{(n)} ,
\end{aligned}
\end{equation}
where \(\mathbf{p}_{i,\mathrm{inv}}^{(n)}\) denotes the probe-invariant block, which is expected to preserve material-discriminative information, and \(\mathbf{p}_{i,\mathrm{sen}}^{(n)}\) denotes the probe-sensitive residual block, which captures local variations induced by context probes.

During training, the context observation network in the Probe-Conditioned Quotient Adapter learns to adaptively allocate material-discriminative information into the invariant coordinates for each sample. We then discard the local subspace that is sensitive to context probes and map the remaining representation back to the original embedding space:
\begin{equation}
    \mathbf{\hat{z}}_i
    =
    \mathbf{Q}_i^{\top}
    \left[
    \mathbf{p}_{i,\mathrm{inv}}^{(0)};
    \mathbf{0}
    \right].
\end{equation}
Equivalently, in the local quotient coordinate system, we retain the probe-invariant block while suppressing or zeroing out the probe-sensitive block, and then recover the representation in the original embedding space through the inverse orthogonal transformation \(\mathbf{Q}_i^{\top}\).

\textbf{Probe-Invariant Consistency Loss.}
The Probe-Invariant Consistency Loss is the key objective for training the Probe-Conditioned Quotient Adapter. It constrains only the invariant block to remain stable across Context-Probing Intervention, while allowing the sensitive block to absorb probe-induced residual variations. To this end, we define the loss as
\begin{equation}
    \mathcal{L}_{\mathrm{PIC}}
    =
    \frac{1}{N_\text{cp}}
    \sum_{n=1}^{N_\text{cp}}
    \left\|
    \mathbf{p}_{i,\mathrm{inv}}^{(0)}
    -
    \mathrm{sg}\left(\mathbf{p}_{i,\mathrm{inv}}^{(n)}\right)
    \right\|_2^2,
\end{equation}
where \(\mathrm{sg}(\cdot)\) denotes the stop-gradient operation. Since the Householder-based transformation is orthogonal, the squared Euclidean distance provides a geometry-preserving consistency measure in the transformed subspace. Notably, this loss does not explicitly constrain the probe-sensitive residual block. Instead, it only requires the probe-invariant blocks of the original sample and its context probes to remain consistent. This design is motivated by the presence of the main classification objective. Under their joint optimization, the Probe-Conditioned Quotient Adapter is encouraged to preserve material-discriminative information in the invariant block, while pushing probe-sensitive variations into the sensitive block in the transformed coordinate space.

\subsection{Probe-Stability Prototype Calibration}

In the incremental stage, each support sample is valuable, but not all support samples are equally reliable. If a support sample exhibits large fluctuations under Context-Probing Intervention, it indicates that the sample is highly sensitive to acquisition context. Directly using such a sample to construct the prototype may therefore introduce context-induced prototype contamination. To address this issue, Probe-Stability Prototype Calibration estimates the relative reliability of few-shot support samples by measuring the uncertainty induced by context probes. PSPC follows Context-Probing Intervention to construct context probes for each support sample:
\begin{equation}
    \left\{
    \tilde{\mathbf{M}}_i^{(n)}
    \right\}_{n=1}^{N_{\mathrm{ps}}}
    =
    T_{\mathrm{cp}}(\mathbf{M}_i),
\end{equation}
where \(N_{\mathrm{ps}}\) denotes the number of context probes generated for each sample. It is worth noting that, in the practical implementation, we use only random view perturbation to construct these context probes for a fair comparison.

After passing the generated probes through the embedding learner, we obtain \(N_{\mathrm{ps}}\) probe embeddings together with the original embedding:
\begin{equation}
\begin{aligned}
    \mathbf{z}_i^{(n)} = f_\theta\left(\tilde{\mathbf{M}}_i^{(n)}\right),
    \quad n=1,&\ldots,N_{\mathrm{ps}},\\
    \mathbf{z}_i^{(0)} = f_\theta(\mathbf{M}_i).
\end{aligned}
\end{equation}
PCQA further maps these embeddings to \(\hat{\mathbf{z}}_i^{(n)}\) and \(\hat{\mathbf{z}}_i^{(0)}\), based on which we compute the probe-stability uncertainty:
\begin{equation}
    U_i
    =
    \frac{1}{N_{\mathrm{ps}}}
    \sum_{n=1}^{N_{\mathrm{ps}}}
    \left\|
    \hat{\mathbf{z}}_i^{(n)} - \hat{\mathbf{z}}_i^{(0)}
    \right\|_2^2 .
\end{equation}
 A larger \(U_i\) indicates that the support sample is unstable under context probes, whereas a smaller \(U_i\) suggests that the sample remains stable under Context-Probing Intervention and is therefore more suitable for representing the material class. Based on this observation, Probe-Stability Prototype Calibration further computes a reliability weight for each support sample in the support set \(S_y\) of a novel class \(y\):
\begin{equation}
    \alpha_{i}
    =
    \frac{
    \exp(-\beta U_i)
    }{
    \sum_{j \in S_y}
    \exp(-\beta U_j)
    },
\end{equation}
where \(\beta\) is a context scaling factor, which is set to \(8\) in this work. The reliability weights are applied to the mean support embeddings \(\bar{\mathbf{z}}_i\), yielding the calibrated prototype mean:
\begin{equation}
    \bm\mu^{\mathrm{pspc}}_{y}
    =
    \sum_{i \in S_y}
    \alpha_{i} \bar{\mathbf{z}}_i .
\end{equation}

This procedure converts the representation fluctuation measured by Context-Probing Intervention into a reliability estimate for prototype construction. As a result, support samples that are highly sensitive to context probes contribute less to the novel class prototype, while more stable samples receive larger weights. This design is particularly useful in the incremental stage, where the support set is extremely limited and may be affected by previously unseen acquisition contexts.

\subsection{Overall Loss Function}

We optimize different objectives in the base and incremental sessions of FSCIL. Following PITS-SC~\cite{li2025fewaudio}, the base session is divided into a pretraining stage, and a full-training stage which incorporates pseudo-incremental training. During the base session, we jointly train the embedding learner and PCQA using the cross-entropy classification loss  $\mathcal{L}_{\text{CE}}$ and the CoP-FSCIL specific Probe-Invariant Consistency Loss \(\mathcal{L}_{\text{PIC}}\). In incremental sessions, we apply classification supervision to newly introduced classes and regularize the classifier with stored class-mean embeddings of previously learned classes to mitigate forgetting. Concretely, we optimize the following losses. For the base session, the objective is defined as
\begin{equation}
\mathcal{L}_{\text{Base}} = \mathcal{L}_{\text{CE}} + \lambda \mathcal{L}_{\text{PIC}}.
\end{equation}
Here, \(\lambda\) is the loss-balancing coefficient and is set to (0.1) in all experiments. For incremental sessions, the objective is defined as
\begin{equation} 
\mathcal{L}_{\text{Inc}} = \mathcal{L}_{\text{New}} + \mathcal{L}_{\text{Old}},
\end{equation}
where the loss for newly introduced classes $\mathcal{Y}_{\ell}$ is
\begin{equation} 
\mathcal{L}_{New} = -\log \left( \frac{e^{\cos \left( \hat{\mathbf{z}}, \tilde{\bm{\mu}}_y^\text{pspc} \right)}}{\sum_{j \in \mathcal{Y}_{\ell}} e^{\cos \left( \hat{\mathbf{z}}, \tilde{\bm\mu}_j^\text{pspc} \right)}} \right),
\end{equation}
and the loss for previously learned classes $\mathcal{Y}_{< \ell}$ is
\begin{equation} 
\mathcal{L}_{Old} = -\log \left( \frac{e^{\cos \left( \bm\mu, \tilde{\bm\mu}_{y'}^\text{pspc} \right)}}{\sum_{j \in \mathcal{Y}_{< \ell}} e^{\cos \left( \bm\mu, \tilde{\bm\mu}_j^\text{pspc} \right)}} \right).
\end{equation}
In $\mathcal{L}_{Old}$, $\bm\mu$ denotes stored class-mean embedding of an old class indexed by $y'$, and the index $j$ ranges over class labels.

\section{Experiments}
\label{Experiments and Discussions}
\subsection{Datasets}
\begin{figure}[ht]
  \centering
    \centerline{\includegraphics[width=0.9\columnwidth]{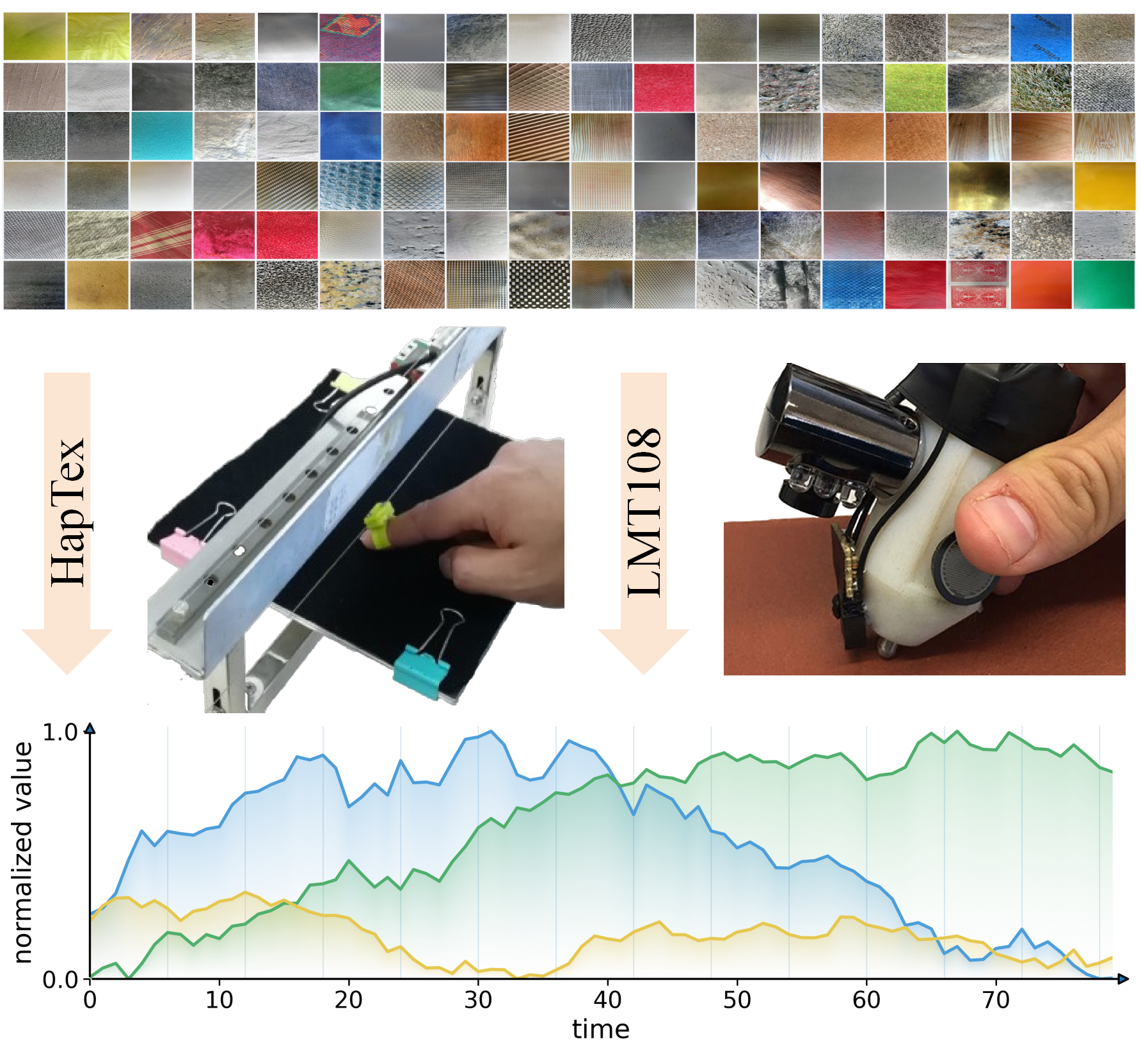}}
    \caption{Photographs show example materials from LMT108. HapTex groups materials into coarse categories ({\it e.g.}, velvet, leather, chiffon, wool, nylon, polyester, linen, and silk). LMT108 defines nine coarse categories, including meshes, stones, blank glossy surfaces, wood types, rubbers, fibers, foams, foils and papers, and textiles and fabrics. We apply dataset-specific normalization for numerical stability and ease of use. Context-dependent variations persist after this preprocessing, motivating CoP-FSCIL.}
    \label{haptex-lmt}
\end{figure}

\textit{1) Main Datasets:} We construct two datasets to evaluate tactile FSCIL from the raw HapTex~\cite{haptex} and LMT108~\cite{lmt108} data. Both are representative numerical--tactile benchmarks and are aligned with the IEEE~1918.1.1 standard format~\cite{8605315} in practice. The acquisition setups are shown in \cref{haptex-lmt}. We slice the raw sequences into short segments and summarize the configurations in \cref{tab:dataset_details}. To reflect short contacts in practical scenarios, we use slice lengths of 0.5~s and 0.8~s for HapTex and LMT108. Each segment is annotated with both coarse and fine-grained labels. Unless otherwise stated, we use only the fine-grained labels in this work throughout.

\begin{table}[ht]
\centering
\caption{Preprocessed dataset details.}
\label{tab:dataset_details}
\renewcommand{\arraystretch}{1.05}
\setlength{\tabcolsep}{0pt}
\footnotesize
\begin{tabular*}{\columnwidth}{@{\extracolsep{\fill}}lcccc@{}}
\toprule
Dataset &
\makecell{Sampling\\Rate (Hz)} &
\makecell{Sample\\Length (s)} &
\makecell{Category\\(Coarse/Fine)} &
\makecell{Num./\\Category} \\
\midrule
HapTex & 1K  & 0.5 & 10/120 & 100 \\
LMT108 & 10K & 0.8 & 9/108  & 120 \\
\bottomrule
\end{tabular*}
\end{table}

HapTex is collected with a TexRecorder system that combines a force sensor (normal and friction forces) and a high-resolution grating displacement sensor (sliding displacement). It contains 120 material classes with time-series measurements of friction force, normal force, friction coefficient, displacement, and velocity. In addition, HapTex provides scanned texture images for each material, which enables appearance-based modeling in some cases.

\begin{table}[!t]
    \centering
    \caption{Log-Mel spectrogram settings and the corresponding physical resolutions for tactile and audio datasets.}
    \label{tab:logmel_params_all}
    \renewcommand{\arraystretch}{1.08}
    \setlength{\tabcolsep}{0pt}
    \footnotesize
    \begin{tabular*}{\columnwidth}{@{\extracolsep{\fill}}lcccc@{}}
        \hline
        Parameter & HapTex & LMT108 & LS-100 & NS-100 \\
        \hline
        Modality & Tactile & Tactile & Audio & Audio \\
        Sampling rate (kHz) & $1$ & $10$ & $16$ & $16$ \\
        FFT size & $128$ & $128$ & $400$ & $2048$ \\
        Hop length (samples) & $16$ & $16$ & $180$ & $1024$ \\
        Window function & Hann & Hann & Hann & Hann \\
        Number of Mel bins & $64$ & $64$ & $128$ & $128$ \\
        \hline
    \end{tabular*}
\end{table}

LMT108 is collected using a handheld Texplorer that supports multimodal tactile sensing. The probe uses a hemispherical stainless-steel tip and integrates a triaxial accelerometer, microphone, infrared reflectance sensor, camera, and dual force-sensitive resistors. Tactile-related signals are captured by an NI DAQ at 10~kHz. The dataset covers 108 material classes and provides synchronized multimodal observations, including acceleration, friction, reflectance, audio, metal detection, and images at high resolution.

\textit{2) Extended Datasets:} Beyond tactile datasets, we further evaluate CoP-FSCIL on two commonly used audio few-shot class-incremental benchmarks, LS-100\cite{arp} and NS-100\cite{arp}, to examine its effectiveness on other modalities with similar spectro-temporal characteristics. LS-100 and NS-100 are constructed from LibriSpeech~\cite{ls} and NSynth~\cite{ns}, respectively, with NSynth being widely used in audio classification studies. Specifically, LibriSpeech is a large-scale audiobook corpus containing approximately 1,000 hours of read English speech from 2,484 speakers. Its development and test sets are each divided into clean and other subsets. LS-100 processes LibriSpeech into speech samples from 100 distinct speakers. NSynth is a large-scale musical-note audio corpus. It contains 306,043 audio clips, each lasting four seconds and corresponding to a musical note with a unique pitch, timbre, and envelope. The full corpus covers 1,006 instruments, and NS-100 includes 100 instrument classes from this corpus.

\subsection{Implementation Details}
\label{Implementation Details}

\textit{1) Log-Mel Spectrogram Settings:} We convert tactile signals into log-Mel spectrograms by computing Hann-windowed STFT magnitude spectra, projecting them onto a Mel filterbank, and applying logarithmic compression. The same extractor and parameters are used for HapTex and LMT108, except for the sampling rate: $1K$~Hz for HapTex and $10K$~Hz for LMT108. Table~\ref{tab:logmel_params_all} summarizes key parameters. For the extended audio datasets, we retain the original preprocessing protocols to ensure consistency with their standard settings.

\textit{2) Baselines and Evaluation Protocols:}
For the main experiments on tactile FSCIL, we compare CoP-FSCIL with PriViLege~\cite{park2024pre}, Comp-FSCIL~\cite{zou2024compositional}, OrCo~\cite{ahmed2024orco}, ADBS~\cite{li2025adaptive}, LRT~\cite{lrt}, PITS-SC~\cite{li2025fewaudio}, TAPE~\cite{gao2026tape}, and PA-PCT~\cite{li_2026_INTERSPEECH}. We use the official implementations for all compared methods. Since most baselines were designed for image benchmarks, except PITS-SC, TAPE and PA-PCT for audio, we standardize the input format by converting tactile signals into log-Mel spectrograms. The resulting log-Mel input sizes are $5 \times 64 \times 24$ for HapTex and $3 \times 64 \times 493$ for LMT108. PriViLege and LRT further resize the inputs to $3 \times 224 \times 224$, which should be considered when comparing computational costs. For extended audio FSCIL experiments, we compare with ARP~\cite{arp}, LDC~\cite{ldc}, DSN~\cite{dsn}, PAN~\cite{pan}, AMFO~\cite{amfo}, PITS-SC~\cite{li2025fewaudio}, TAPE~\cite{gao2026tape}, and PA-PCT~\cite{li_2026_INTERSPEECH}, following the evaluation setting of PITS-SC~\cite{li2025fewaudio}. The resulting log-Mel input sizes are $3 \times 128 \times 63$ for LS-100 and $3 \times 128 \times 178$ for NS-100. All experiments follow the standard 5-way 5-shot protocol.

\begin{table*}[!t]
\centering
\caption{Comparison on the HapTex dataset. \best{Red} and \second{blue} denote the best and second-best results, respectively. All values are reported in percentage.}
\label{tab:session_acc}
\renewcommand{\arraystretch}{1.14} \setlength{\tabcolsep}{3.0pt}
\small
\begin{tabular}{@{}l@{\hspace{8pt}}l@{\hspace{8pt}}*{10}{C{0.78cm}}@{\hspace{8pt}}C{0.78cm}C{0.78cm}C{0.78cm}@{}}
\toprule
\multirow{2}{*}{Method} &
\multirow{2}{*}{Pub. Year} &
\multicolumn{10}{c}{Accuracy in each session $\uparrow$} &
\multirow{2}{*}{PD$\downarrow$} &
\multirow{2}{*}{AA$\uparrow$} &
\multirow{2}{*}{ADR$\downarrow$} \\
\cmidrule(lr){3-12}
& & 0 & 1 & 2 & 3 & 4 & 5 & 6 & 7 & 8 & 9 \\
\midrule

ADBS\cite{li2025adaptive} 
& AAAI25 
& 92.95 & 83.83 & 72.58 & 75.13 & 68.77 & 62.58 & 65.09 & 62.87 & 61.02 & 58.95 
& 34.00 & 70.38 & 4.77 \\

Comp-FSCIL\cite{zou2024compositional} 
& ICML24 
& 93.79 & 88.75 & 83.91 & 79.71 & 76.06 & 72.46 & 69.98 & 66.95 & 64.42 & 62.60 
& 31.19 & 75.86 & 4.39 \\

PriViLege\cite{park2024pre} 
& CVPR24 
& 83.79 & 73.38 & 67.15 & 65.20 & 62.59 & 60.22 & 58.23 & 56.29 & 54.56 & 53.18 
& 30.61 & 63.46 & 4.87 \\

TAPE\cite{gao2026tape} 
& CVPR26
& 88.90 & 82.66 & 77.90 & 75.73 & 72.46 & 69.42 & 65.95 & 62.15 & 60.17 & 59.12 & 29.78 & 71.45 & 4.42 \\

LRT\cite{lrt} 
& TPAMI25
& 90.15 & 86.05 & 84.67 & 80.91 & 75.85 & 72.82 & 68.70 & 66.20 & 62.78 & 60.68 
& 29.47 & 74.88 & 4.29 \\

PITS-SC\cite{li2025fewaudio} 
& TASLP25 
& \second{96.54} & \second{91.69} & \second{87.66} & \second{85.68} 
& \second{81.56} & 77.76 
& 75.83 & 74.04 & 71.13 & 69.72 
& 26.82 & 81.16 & 3.54 \\

PA-PCT\cite{li_2026_INTERSPEECH} 
& Interspeech26
& 95.68 & 90.80 & 86.86 & 84.74 & 81.00 & \second{79.14} & \second{78.34} & \second{75.12} & \second{72.34} & \second{70.82} & 24.86 & \second{81.48} & 3.28 \\

OrCo\cite{ahmed2024orco} 
& CVPR24 
& 94.72 & 90.78 & 86.80 & 84.62 & 80.34 & 77.46 
& 75.60 & 73.76 & 71.88 & 70.73
& \second{23.99} & 80.67 & \second{3.19} \\

CoP-FSCIL 
& Ours 
& \best{97.05} & \best{93.44} & \best{89.91} & \best{88.13} 
& \best{84.56} & \best{81.06} & \best{79.15} & \best{77.39} 
& \best{75.18} & \best{74.15} 
& \best{22.90} & \best{84.00} & \best{2.94} \\

\bottomrule
\end{tabular}
\end{table*}

\begin{table*}[!t]
\centering
\caption{Comparison on the LMT108 dataset.}
\label{tab2:session_acc}
\renewcommand{\arraystretch}{1.14} \setlength{\tabcolsep}{3.0pt}
\small
\begin{tabular}{@{}l@{\hspace{8pt}}l@{\hspace{8pt}}*{10}{C{0.78cm}}@{\hspace{8pt}}C{0.78cm}C{0.78cm}C{0.78cm}@{}}
\toprule
\multirow{2}{*}{Method} &
\multirow{2}{*}{Pub. Year} &
\multicolumn{10}{c}{Accuracy in each session $\uparrow$} &
\multirow{2}{*}{PD$\downarrow$} &
\multirow{2}{*}{AA$\uparrow$} &
\multirow{2}{*}{ADR$\downarrow$} \\
\cmidrule(lr){3-12}
& & 0 & 1 & 2 & 3 & 4 & 5 & 6 & 7 & 8 & 9 \\
\midrule

ADBS\cite{li2025adaptive} 
& AAAI25
& \best{55.07} & \best{48.27} & \second{42.29} & 33.41 & 28.15 & 21.72 & 16.23 & 12.98 & 10.87 & 8.19
& 46.88 & 27.69 & 18.94 \\

Comp-FSCIL\cite{zou2024compositional} 
& ICML24
& 42.54 & 20.47 & 17.37 & 16.28 & 12.81 & 11.40 & 10.80 & 10.20 & 10.97 & 9.30
& 33.24 & 16.21 & 13.79 \\

PriViLege\cite{park2024pre} 
& CVPR24
& 41.95 & 22.23 & 20.75 & 19.55 & 18.53 & 17.46 & 16.61 & 15.85 & 15.03 & 14.34
& 27.61 & 20.23 & 9.96 \\

TAPE\cite{gao2026tape} 
& CVPR26
& 38.81 & 30.71 & 25.25 & 20.21 & 18.31 & 17.40 & 15.85 & 14.92 & 13.87 & 13.02 & 25.79 & 20.84 & 11.21 \\

LRT\cite{lrt} 
& TPAMI25
& 40.59 & 36.33 & 32.58 & 29.33 & 27.04 & 24.48 & 22.21 & 20.57 & 18.71 & 17.04 
& 23.55 & 26.89 & 9.19 \\
                                             
PITS-SC\cite{li2025fewaudio} 
& TASLP25
& 43.86 & 39.56 & 36.87 & 34.31 & 33.02 & 31.39 & 29.73 & 28.08 & 26.74 & 23.38
& 20.48 & 32.69 & 6.71 \\

PA-PCT\cite{li_2026_INTERSPEECH} 
& Interspeech26
& 44.21 & 40.80 & 38.94 & \second{36.62} & \second{34.53} & \second{32.70} & \second{30.89} & \second{29.02} & 27.02 & 24.77 & 19.44 & \second{33.95} & 6.23 \\
 	 	 	 	 	 	 	 	 	 	 	 	 	 
OrCo\cite{ahmed2024orco} 
& CVPR24
& 44.13 & 39.27 & 37.40 & 35.24 & 33.35
& 31.29 & 29.73 & 28.55 & \second{27.04} & \second{25.56}
& \best{18.57} & 33.16 & \second{5.87} \\
                                                 
CoP-FSCIL 
& Ours
& \second{47.34} & \second{47.03} & \best{43.49} & \best{40.25} & \best{37.59}
& \best{35.05} & \best{33.21} & \best{31.41} & \best{29.97} & \best{28.06}
& \second{19.28} & \best{37.34} & \best{5.63} \\
 	 	 	 	 	 	 	 	 	 	 	 	 
\bottomrule
\end{tabular}
\end{table*}

\begin{figure*}[!ht]
  \centering
  \includegraphics[width=\textwidth]{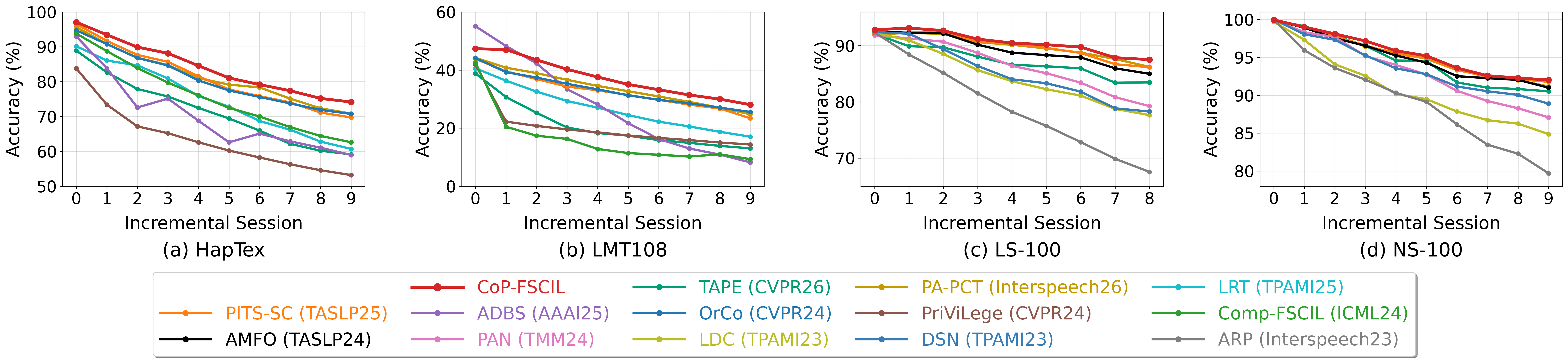}
 \caption{Performance comparison results. (a) HapTex and (b) LMT108 are tactile datasets, whereas (c) LS-100 and (d) NS-100 are audio datasets. CoP-FSCIL achieves the strongest overall performance across datasets, obtaining the best results on most metrics and competitive results on the remaining ones.}
  \label{caotu3}
\end{figure*}

\begin{figure}[!t]
  \centering
    \centerline{\includegraphics[width=\columnwidth]{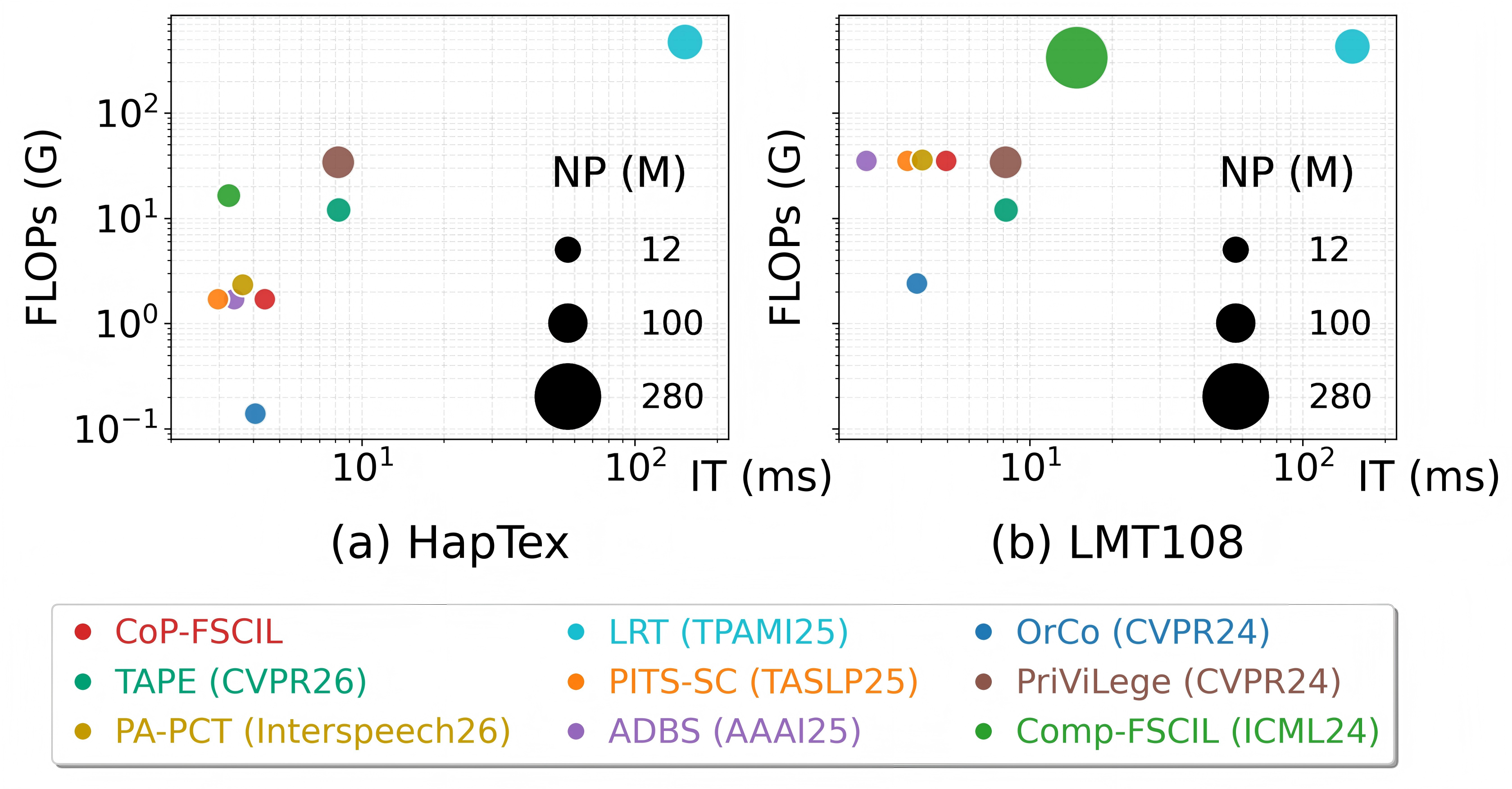}}
    \caption{Computational efficiency comparisons on HapTex and LMT108.
}
    \label{fig:efficiency_comparison}
\end{figure}

\textit{3) Evaluation Metrics:} We evaluate all methods using Average Accuracy ($\text{AA}$), Performance Drop ($\text{PD}$), and Average Drop Rate ($\text{ADR}$). Let $\text{Acc}_{\ell}$ denote the accuracy after session $\ell$, evaluated on the cumulative label space containing all classes observed up to $\ell$. Intuitively, $\text{AA}$ measures overall performance averaged across sessions, $\text{PD}$ reflects end-to-end degradation from base to final, and $\text{ADR}$ captures the average relative drop between successive sessions:
\begin{equation}
 \begin{aligned}
  &\text{AA} = \frac{1}{S+1}\sum_{\ell=0}^{S} \text{Acc}_{\ell},\\
  &\text{PD} = \text{Acc}_0 - \text{Acc}_S,\\
  &\text{ADR} = \frac{1}{S}\sum_{\ell=0}^{S-1}\frac{\text{Acc}_{\ell}-\text{Acc}_{\ell+1}}{\text{Acc}_{\ell}}.
 \end{aligned}
\end{equation}
We use $\text{AA}$ as the primary metric since it summarizes overall performance across sessions in a single scalar, while $\text{PD}$ and $\text{ADR}$ complement it by quantifying performance degradation and stability over incremental updates. 

In addition, we evaluate methods using Floating Point Operations (FLOPs), Number of Parameters (NP), and Inference Time (IT) to compare their computational costs.

\begin{table*}[!t]
\centering
\caption{Comparison on the LS-100 dataset.}
\label{tab:session_accLS}
\renewcommand{\arraystretch}{1.14} \setlength{\tabcolsep}{3.0pt}
\small

\begin{tabular}{@{}l@{\hspace{8pt}}l@{\hspace{8pt}}*{9}{C{0.845cm}}@{\hspace{8pt}}C{0.845cm}C{0.845cm}C{0.845cm}@{}}
\toprule
\multirow{2}{*}{Method} &
\multirow{2}{*}{Pub. Year} &
\multicolumn{9}{c}{Accuracy in each session $\uparrow$} &
\multirow{2}{*}{PD$\downarrow$} &
\multirow{2}{*}{AA$\uparrow$} &
\multirow{2}{*}{ADR$\downarrow$} \\
\cmidrule(lr){3-11}
& & 0 & 1 & 2 & 3 & 4 & 5 & 6 & 7 & 8 \\
\midrule

ARP\cite{arp} 
& Interspeech23
& 92.35 & 88.43 & 85.16 & 81.53 & 78.22 & 75.73 & 72.84 & 69.87 & 67.54
& 24.81 & 79.07 & 3.83 \\

LDC\cite{ldc} 
& TPAMI23
& 92.28 & 90.95 & 88.53 & 85.64 & 83.67 & 82.25 & 81.11 & 78.75 & 77.64
& 14.64 & 84.54 & 2.13 \\

DSN\cite{dsn} 
& TPAMI23
& 92.43 & 92.12 & 89.37 & 86.39 & 84.02 & 83.31 & 81.82 & 78.85 & 78.28
& 14.15 & 85.18 & 2.05 \\
 	 	 	 	 		 	 	 	 	 	 
PAN\cite{pan} 
& TMM24
& 91.83 & 91.29 & 90.70 & 88.73 & 86.42 & 85.09 & 83.41 & 80.84 & 79.24
& 12.59 & 86.39 & 1.82 \\

TAPE\cite{gao2026tape} 
& CVPR26
& 92.10 & 89.91	& 89.71	& 88.05	& 86.60 & 86.33 & 85.96	& 83.41	& 83.47	& 8.63 & 87.28 &	1.22 \\ 

AMFO\cite{amfo} 
& TASLP24
& \second{92.63} & 92.25 & 92.20 & 90.17 & 88.75 & 88.32 & 87.90 & 85.97 & 84.99
& 7.64 & 89.24 & 1.07 \\

PITS-SC\cite{li2025fewaudio} 
& TASLP25
& 92.03 & \second{92.41} & \second{92.31} & \second{90.77}
& \second{90.36} & \second{89.60} & 88.70
& 86.70 & 86.17
& 5.86 & 89.89 & 0.82 \\

PA-PCT\cite{li_2026_INTERSPEECH} 
& Interspeech26
& 92.01 & 92.27 & 92.10 & 90.63 & 90.16 & 89.50 & \second{88.76} & \second{87.63} & \second{86.22} & \second{5.79}  & \second{89.92} & \second{0.81} \\

CoP-FSCIL 
& Ours
& \best{92.83} & \best{93.13} & \best{92.68}
& \best{91.17} & \best{90.47} & \best{90.17}
& \best{89.75} & \best{87.86} & \best{87.52}
& \best{5.31} & \best{90.62} & \best{0.73} \\

\bottomrule
\end{tabular}
\end{table*}

\begin{table*}[!t]
\centering
\caption{Comparison on the NS-100 dataset. }
\label{tab:session_acc_2NS}
\renewcommand{\arraystretch}{1.14} \setlength{\tabcolsep}{3.0pt}
\small

\begin{tabular}{@{}l@{\hspace{8pt}}l@{\hspace{8pt}}*{10}{C{0.78cm}}@{\hspace{8pt}}C{0.78cm}C{0.78cm}C{0.78cm}@{}}
\toprule
\multirow{2}{*}{Method} &
\multirow{2}{*}{Pub. Year} &
\multicolumn{10}{c}{Accuracy in each session $\uparrow$} &
\multirow{2}{*}{PD$\downarrow$} &
\multirow{2}{*}{AA$\uparrow$} &
\multirow{2}{*}{ADR$\downarrow$} \\
\cmidrule(lr){3-12}
& & 0 & 1 & 2 & 3 & 4 & 5 & 6 & 7 & 8 & 9 \\
\midrule

ARP\cite{arp} 
& Interspeech23
& 99.96 & 95.95 & 93.60 & 92.06 & 90.32 & 89.14 & 86.16 & 83.47 & 82.28 & 79.69
& 20.27 & 89.26 & 2.48 \\

LDC\cite{ldc} 
& TPAMI23
& 99.71 & 97.32 & 94.09 & 92.57 & 90.16 & 89.49 & 87.85 & 86.71 & 86.25 & 84.87
& 14.84 & 90.90 & 1.77 \\

PAN\cite{pan} 
& TMM24
& 99.98 & 98.32 & 97.69 & 95.18 & 94.00 & 92.73 & 90.59 & 89.24 & 88.28 & 87.08
& 12.90 & 93.31 & 1.52 \\

DSN\cite{dsn} 
& TPAMI23
& \best{100.00} & 98.05 & 97.32 & 95.27 & 93.56 & 92.78 & 91.16 & 90.53 & 90.05 & 88.89
& 11.11 & 93.76 & 1.30 \\

TAPE\cite{gao2026tape} 
& CVPR26
& 99.91 & \best{99.12} & 97.86 & 96.53 & 94.59 & 94.53 & 91.69 & 91.01 & 90.82 & 90.52 & 9.39 & 94.66 & 1.09 \\

AMFO\cite{amfo} 
& TASLP24
& 99.95 & 98.92 & 97.54 & 96.54 & 95.25 & 94.31 & 92.52 & 92.26 & 92.03 & 91.00
& 8.95 & 95.03 & 1.04 \\

PA-PCT\cite{li_2026_INTERSPEECH} 
& Interspeech26
& \second{99.98} & 98.90 & \second{97.96} & 96.42 & \second{95.64} & 94.81 & 93.32 & \second{92.52} & 91.99 & 91.16 & 8.82 & 95.27 & 1.02 \\

PITS-SC\cite{li2025fewaudio} 
& TASLP25
& 99.98 & 99.00 & 97.77
& \second{96.58} & 95.43 & \second{94.93}
& \second{93.39} & 92.42 & \second{92.13}
& \second{91.77}
& \second{8.21} & \second{95.34} & \second{0.95} \\

CoP-FSCIL 
& Ours
& 99.95 & \second{99.01} & \best{98.10}
& \best{97.17} & \best{95.88} & \best{95.21}
& \best{93.60} & \best{92.58} & \best{92.28}
& \best{92.02}
& \best{7.93} & \best{95.58} & \best{0.91} \\

\bottomrule
\end{tabular}
\end{table*}

\textit{4) Optimization:} We train all models with SGD. For the main tactile experiments, learning rates for pretraining, full base training, and incremental training are 0.1, 0.01, and 0.1, respectively, with corresponding epoch numbers of 100, 10, and 200. The learning rate is decayed by a factor of 0.5 every 40 epochs. We set \(\lambda\) to 0.1. For PSPC, \(N_{\mathrm{ps}}\) is set to 100 on HapTex and 10 on LMT108. These settings are fixed across all runs and methods on the main tactile datasets for comparison. For the audio datasets, only the epoch numbers are adjusted: 10, 10, and 100 on LS-100, and 40, 40, and 200 on NS-100, for pretraining, full base training, and incremental training. 


\subsection{Comparative Results}

\textit{1) Tactile FSCIL Performance:}
The results on HapTex and LMT108 are summarized in Table~\ref{tab:session_acc}, Table~\ref{tab2:session_acc}, and Fig.~\ref{caotu3} (a) (b). CoP-FSCIL achieves the best overall results on both datasets, except that its PD ranks second on LMT108. On HapTex, CoP-FSCIL obtains an AA of 84.00\%, exceeding the second-best   PA-PCT by 2.52 percentage points. On LMT108, CoP-FSCIL improves AA from the second-best result of 33.95\% to 37.34\%, corresponding to a gain of 3.39 percentage points. The competitive PD and lower ADR further show that CoP-FSCIL improves recognition accuracy while maintaining stronger stability across incremental sessions.

\begin{table*}[!t]
\centering
\caption{Ablation study on HapTex dataset.}
\label{tab:ablation_progressive_full}
\setlength{\tabcolsep}{2.6pt}
\renewcommand{\arraystretch}{1.12}
\small
\begin{tabular}{l l *{10}{c} c c c}
\toprule
\multirow{2}{*}{Variant} &
\multirow{2}{*}{Probe-related design} &
\multicolumn{10}{c}{Accuracy in each session $\uparrow$} &
\multirow{2}{*}{PD$\downarrow$} &
\multirow{2}{*}{AA$\uparrow$} &
\multirow{2}{*}{ADR$\downarrow$} \\
\cmidrule(lr){3-12}
& & 0 & 1 & 2 & 3 & 4 & 5 & 6 & 7 & 8 & 9 & & & \\
\midrule

Baseline
& No context probe
& 96.54 & 91.69 & 87.66 & 85.68 & 81.56 & 77.76 & 75.83 & 74.04 & 71.13 & 69.72
& 26.82 & 81.16 & 3.54 \\

+CPI
& Probe construction
& \textbf{\textcolor{blue}{97.52}} & 93.16 & 89.06 & 87.13 & 83.17 & 79.56 & 77.68 & 75.54 & 72.90 & 71.50
& 26.02 & 82.72 & 3.38 \\

+CPI+PSPC
& Probe-stability prototype
& \textbf{\textcolor{red}{97.55}} & 93.24 & 89.17 & 87.30 & 83.35 & 79.75 & 77.90 & 75.79 & 73.13 & 71.71
& 25.84 & 82.89 & 3.36 \\

+CPI+PCQA
& Quotient adaptation
& 97.02 
& \textbf{\textcolor{blue}{93.44}} 
& \textbf{\textcolor{blue}{89.87}} 
& \textbf{\textcolor{blue}{88.05}} 
& \textbf{\textcolor{blue}{84.43}} 
& \textbf{\textcolor{blue}{80.96}} 
& \textbf{\textcolor{blue}{79.03}} 
& \textbf{\textcolor{blue}{77.25}} 
& \textbf{\textcolor{blue}{75.11}} 
& \textbf{\textcolor{blue}{74.04}}
& \textbf{\textcolor{blue}{22.98}} 
& \textbf{\textcolor{blue}{83.92}} 
& \textbf{\textcolor{blue}{2.95}} \\

\textbf{CoP-FSCIL}
& Full framework
& 97.05 
& \textbf{\textcolor{red}{93.44}} 
& \textbf{\textcolor{red}{89.91}} 
& \textbf{\textcolor{red}{88.13}} 
& \textbf{\textcolor{red}{84.56}} 
& \textbf{\textcolor{red}{81.06}} 
& \textbf{\textcolor{red}{79.15}} 
& \textbf{\textcolor{red}{77.39}} 
& \textbf{\textcolor{red}{75.18}} 
& \textbf{\textcolor{red}{74.15}}
& \textbf{\textcolor{red}{22.90}} 
& \textbf{\textcolor{red}{84.00}} 
& \textbf{\textcolor{red}{2.94}} \\

\bottomrule
\end{tabular}
\end{table*}

\begin{table*}[!t]
\centering
\caption{Ablation study on LMT108 dataset. }
\label{tab:ablation_progressive_lmt108}
\setlength{\tabcolsep}{2.6pt}
\renewcommand{\arraystretch}{1.12}
\small
\begin{tabular}{l l *{10}{c} c c c}
\toprule
\multirow{2}{*}{Variant} &
\multirow{2}{*}{Probe-related design} &
\multicolumn{10}{c}{Accuracy in each session $\uparrow$} &
\multirow{2}{*}{PD$\downarrow$} &
\multirow{2}{*}{AA$\uparrow$} &
\multirow{2}{*}{ADR$\downarrow$} \\
\cmidrule(lr){3-12}
& & 0 & 1 & 2 & 3 & 4 & 5 & 6 & 7 & 8 & 9 & & & \\
\midrule

Baseline
& No context probe
& 43.86 & 39.56 & 36.87 & 34.31 & 33.02 & 31.39 & 29.73 & 28.08 & 26.74 & 23.38
& 20.48 & 32.69 & 6.71 \\

+CPI
& Probe construction
&46.11&41.15&37.6&35.48&33.42&31.64&29.77&28.11&26.84&25.54&20.57&33.57 &6.33 \\

+CPI+PSPC
& Probe-stability prototype
& 46.30 & 41.55 & 38.09 & 35.92 & 33.72 & 31.88 & 30.03 & 28.39 & 26.96 & 25.69 & 20.61 & 33.85 & 6.32 \\

+CPI+PCQA
& Quotient adaptation
& \textbf{\textcolor{red}{47.35}}
& \textbf{\textcolor{blue}{46.95}}
& \textbf{\textcolor{blue}{43.42}}
& \textbf{\textcolor{blue}{40.19}}
& \textbf{\textcolor{blue}{37.54}}
& \textbf{\textcolor{blue}{35.00}}
& \textbf{\textcolor{blue}{33.16}}
& \textbf{\textcolor{blue}{31.36}}
& \textbf{\textcolor{blue}{29.88}}
& \textbf{\textcolor{blue}{27.88}}
& \textbf{\textcolor{blue}{19.47}}
& \textbf{\textcolor{blue}{37.27}}
& \textbf{\textcolor{blue}{5.70}} \\
	 	 	 	 	 	 	 	 	 	 	 	 
\textbf{CoP-FSCIL}
& Full framework
& \second{47.34} & \best{47.03} & \best{43.49} & \best{40.25} & \best{37.59}
& \best{35.05} & \best{33.21} & \best{31.41} & \best{29.97} & \best{28.06}
& \best{19.28} & \best{37.34} & \best{5.63} \\
\bottomrule
\end{tabular}
\end{table*}

\begin{figure}[!t]
  \centering
    \centerline{\includegraphics[width=\columnwidth]{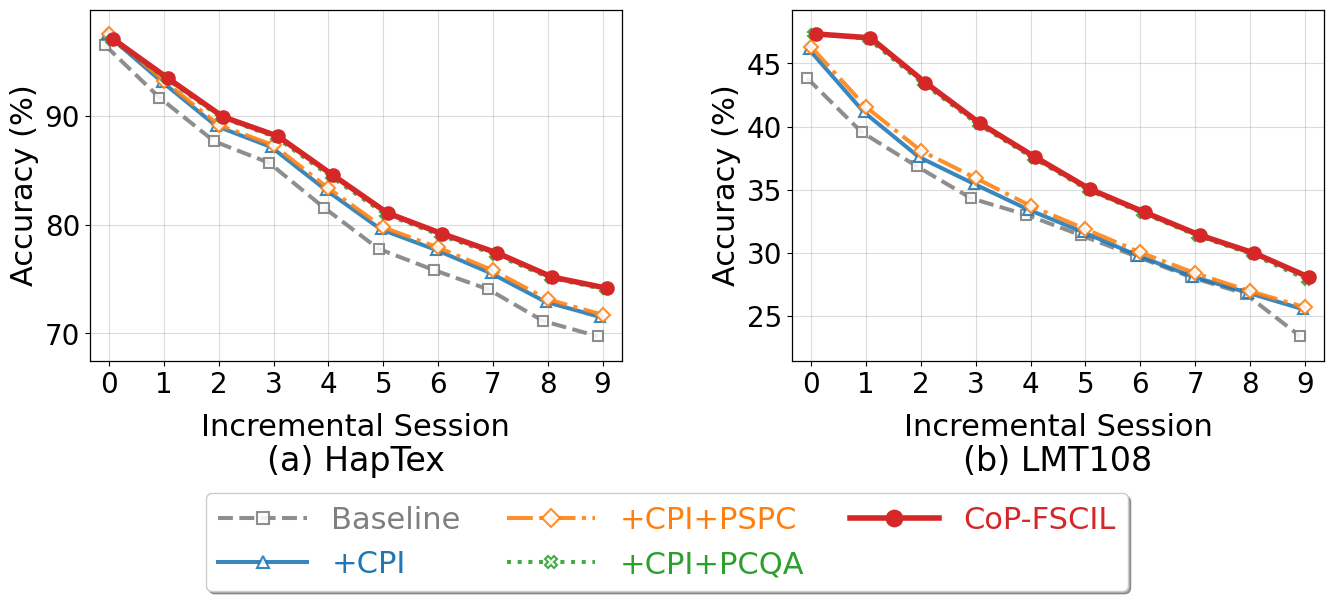}}
    \caption{Performance ablation results on HapTex and LMT108.
}
    \label{caotu4}
\end{figure}

\textit{2) Analysis Across FSCIL Paradigms:}
The comparisons reveal how representative FSCIL paradigms behave under acquisition-context variations. Prompt-based methods, such as PriViLege and LRT, are less effective in tactile scenarios because tactile foundation models and large-scale tactile--text paired data remain limited, making textual prompts difficult to align with log-Mel tactile representations. Boundary-based methods such as ADBS may further amplify spurious context cues when decision boundaries are calibrated without explicitly reducing context dependence, yielding seemingly high base session accuracy but fragile boundaries in later incremental sessions. Replay-based method such as OrCo improves context coverage by storing and replaying real samples, but this benefit comes with a trade-off that requires an explicit memory budget. For audio oriented methods, TAPE relies on the frozen PENGI~\cite{pengi}, whose audio-language prior is not directly transferable to tactile sensing. PITS-SC remains competitive because its pseudo-incremental strategy also enriches context coverage, but it does not separate material-intrinsic information from context-sensitive components. PA-PCT achieves strong performance through its class-adaptive design, which also improves adaptation to context shifts. In contrast, CoP-FSCIL explicitly characterizes context sensitivity through context probes and suppresses context-dependent components during prototype construction, leading to stronger robustness under acquisition-context variations.

\textit{3) Computational Efficiency:}
The efficiency results are reported in Fig.~\ref{fig:efficiency_comparison}. PriViLege and LRT require fixed input sizes; therefore, their inputs are resized to a common resolution across datasets, and the remaining efficiency differences mainly arise from the dataset-specific number of classes. Comp-FSCIL is sensitive to the spatial size of spectrograms, leading to larger cost variations across datasets with different input resolutions. OrCo achieves lower FLOPs mainly because its implementation applies more aggressive downsampling in the backbone. For the other methods, the complexity metrics are generally close. CoP-FSCIL introduces only marginal overhead over PITS-SC, with almost unchanged FLOPs and NP, because PCQA is applied after the embedding learner and is implemented as a lightweight module with few parameters. Overall, CoP-FSCIL incurs only a slight increase in IT while maintaining competitive computational efficiency.

\textit{4) Extension to Audio FSCIL:} The results on LS-100 and NS-100 are summarized in Table~\ref{tab:session_accLS}, Table~\ref{tab:session_acc_2NS}, and Fig.~\ref{caotu3} (c) (d). Interestingly, CoP-FSCIL achieves the strongest overall performance on both audio datasets. For LS-100, where the target label is speaker identity, task-irrelevant factors such as speech content, volume, and temporal segment location may be weakly related or unrelated to the target semantics. These factors can be analogously regarded as acquisition contexts in tactile signals. CoP-FSCIL is well suited to this setting, as it explicitly models and suppresses context-dependent representation components. Similarly, on NS-100, CoP-FSCIL can reduce the influence of weakly task-relevant factors, leading to a representation space that is more aligned with the target semantics. These results further demonstrate the potential generality of CoP-FSCIL beyond tactile FSCIL.

\subsection{Performance Analysis}
\label{Performance Analysis}

\textit{1) Ablation Study:} We conduct progressive ablation studies on HapTex and LMT108 to evaluate each component's contribution, as shown in \cref{tab:ablation_progressive_full}, \cref{tab:ablation_progressive_lmt108}, and \cref{caotu4}. Adding CPI improves AA from 81.16\% to 82.72\% on HapTex and from 32.69\% to 33.57\% on LMT108, indicating that context probes provide useful perturbation cues for learning context-robust representations. When PCQA is further introduced, AA increases to 83.92\% on HapTex and 37.27\% on LMT108, showing that explicitly separating probe-invariant and probe-sensitive components is key to the performance gain. PSPC brings additional but consistent improvements by calibrating prototypes according to probe stability. The full CoP-FSCIL achieves the best overall performance, with AA of 84.00\% on HapTex and 37.34\% on LMT108. These results demonstrate the complementarity of CPI, PCQA, and PSPC.

\begin{figure}[!t]
  \centering
    \centerline{\includegraphics[width=\columnwidth]{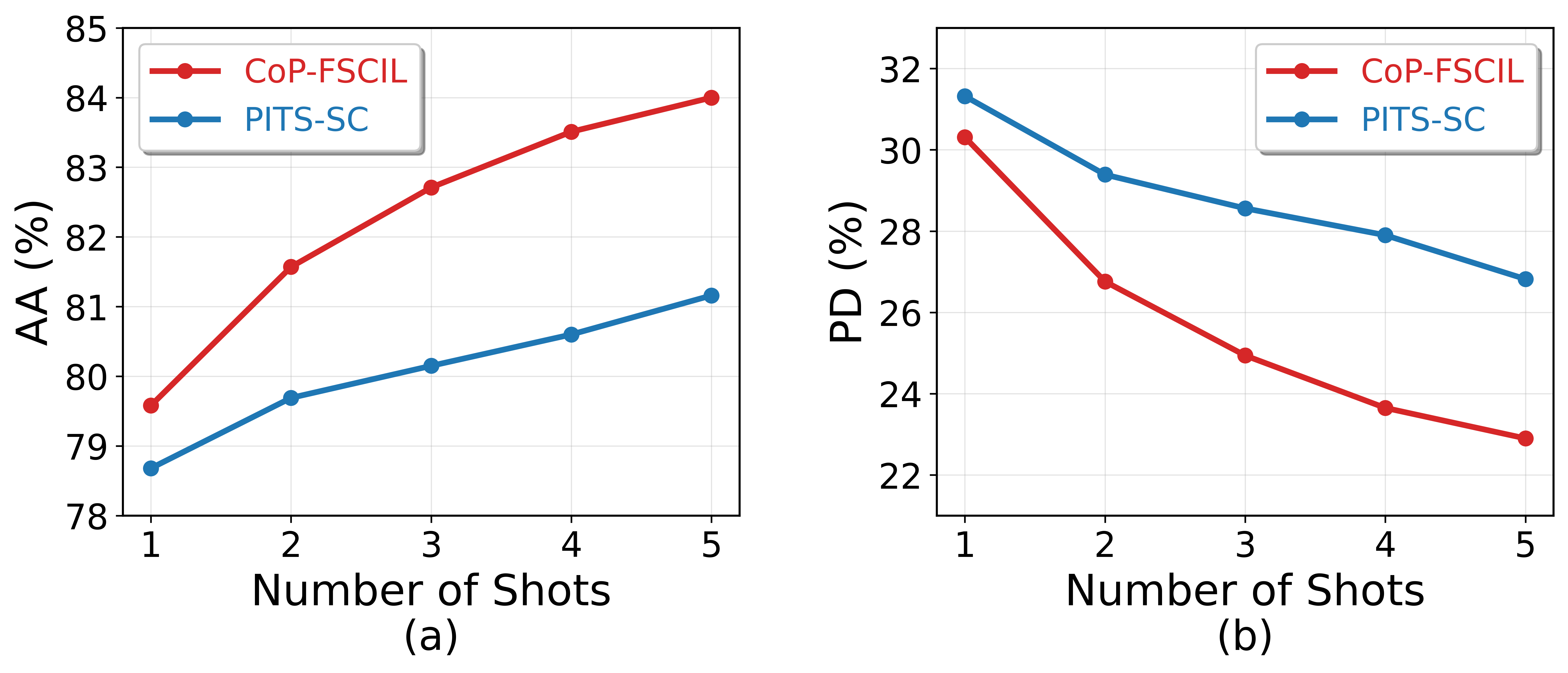}}
    \caption{Average Accuracy and Performance Drop with different number of shots during incremental sessions on HapTex.
}
    \label{fig:xxx}
\end{figure}

\begin{table}[t]
\centering
\caption{Sensitivity analysis of the loss weight \(\lambda\) for \(\mathcal{L}_{\text{PIC}}\) on HapTex.}
\label{tab:lambda_sensitivity}
\setlength{\tabcolsep}{5pt}
\renewcommand{\arraystretch}{1.15}
\begin{tabular*}{\columnwidth}{@{\extracolsep{\fill}}lcccc@{}}
\toprule
$\lambda$ & $\text{Acc}_{0}$$\uparrow$ & PD$\downarrow$ & AA$\uparrow$ & ADR$\downarrow$ \\
\midrule
0        & \second{97.09} & 23.62 & 83.63 & 3.04 \\
1        & \best{97.10} & 24.92 & 83.06 & 3.24 \\
0.1      & 97.05 & \best{22.90} & \best{84.00} & \best{2.94} \\
0.01     & 97.04 & 23.31 & 83.84 & 3.01 \\
0.001    & 97.04 & \second{22.97} & \second{83.92} & \second{2.97} \\
0.0001   & 97.04 & 23.06 & 83.92 & 2.97 \\
\bottomrule
\end{tabular*}
\end{table}

\textit{2) Incremental Learning With Fewer Shots:} Beyond the commonly used \(N\)-way 5-shot setting in incremental learning, we further evaluate CoP-FSCIL under more challenging few-shot regimes ranging from 1-shot to 4-shot. As shown in \cref{fig:xxx}, on the HapTex dataset, CoP-FSCIL achieves higher AA and lower PD than the baseline across all settings from 1-shot to 5-shot. Notably, even under the 2-shot setting, CoP-FSCIL already surpasses the baseline under its standard 5-shot configuration. However, CoP-FSCIL is also more sensitive to the number of shots, exhibiting a more pronounced performance drop as the number of support samples decreases. We attribute this trend to two factors. First, with fewer support samples, PCQA receives insufficient evidence for sample-adaptive coordinate observation, which may lead to increased reliance on context-sensitive representations. Second, the effect of PSPC becomes weakened because probe-stability uncertainty provides less reliable weighting signals for prototype construction, thereby reducing the effectiveness of prototype calibration.

\textit{3) Effect of Loss Constraints:} We conduct experiments with different orders of magnitude of the loss weight \(\lambda\), and report the results in Table~\ref{tab:lambda_sensitivity}. Both increasing and decreasing \(\lambda\) from an appropriate range lead to degraded performance in AA and PD. When \(\lambda\) is set to a large value, such as \(1\), the performance drops substantially and can even become worse than the setting without \(\mathcal{L}_{\text{PIC}}\), i.e., \(\lambda=0\). This is because an overly strong \(\mathcal{L}_{\text{PIC}}\) constraint may interfere with the dominant supervision provided by the classification loss. It is worth noting that \(\mathcal{L}_{\text{PIC}}\) does not necessarily improve the representation ability of the embedding learner on base classes. In some cases, it may even slightly reduce the base session accuracy. However, other metrics, such as PD and ADR, indicate that \(\mathcal{L}_{\text{PIC}}\) helps PCQA mitigate context overfitting in the embedding learner and encourages the model to concentrate its representation capacity on context-invariant features. This explains why an appropriate \(\lambda\) improves incremental robustness even when the gain on base-session accuracy is not evident.

\begin{table}[!t]
\centering
\caption{Effect of pseudo-context statistic transfer strategies in CPI.}
\label{tab:pseudo_context_transfer}
\renewcommand{\arraystretch}{1.12}
\setlength{\tabcolsep}{1.2pt}
\footnotesize
\begin{tabular*}{\columnwidth}{@{\extracolsep{\fill}}lcccc@{}}
\toprule
Strategy 
& $\text{Acc}_{0}\uparrow$ 
& PD$\downarrow$ 
& AA$\uparrow$ 
& ADR$\downarrow$ \\
\midrule
Random selection 
& \best{97.34} & 24.23 & 83.75 & 3.12 \\
Farthest context-statistic selection 
& 97.05 & \best{22.90} & \best{84.00} & \best{2.94} \\
\bottomrule
\end{tabular*}
\end{table}

\begin{table}[!t]
\centering
\caption{Effect of different CPI coverage settings on support and query samples.}
\label{tab:cpi_coverage}
\renewcommand{\arraystretch}{1.10}
\setlength{\tabcolsep}{2.2pt}
\footnotesize
\begin{tabular*}{\columnwidth}{@{\extracolsep{\fill}}cccccccc@{}}
\toprule
\multicolumn{2}{c}{Support} 
& \multicolumn{2}{c}{Query} 
& $\text{Acc}_{0}\uparrow$ 
& PD$\downarrow$ 
& AA$\uparrow$ 
& ADR$\downarrow$ \\
\cmidrule(lr){1-2} \cmidrule(lr){3-4}
$100\%$ & $60\%$ & $100\%$ & $60\%$ & & & & \\
\midrule
$\checkmark$ & $\times$      & $\checkmark$ & $\times$      & 96.90 & \second{23.28} & \second{83.77} & \second{3.00} \\
$\checkmark$ & $\times$      & $\times$      & $\checkmark$ & 96.87 & 24.15 & 83.11 & 3.13 \\
$\times$      & $\checkmark$ & $\checkmark$ & $\times$      & \second{97.05} & \best{22.90} & \best{84.00} & \best{2.94} \\
$\times$      & $\checkmark$ & $\times$      & $\checkmark$ & \best{97.46} & 24.52 & 83.44 & 3.17 \\
\bottomrule
\end{tabular*}
\end{table}

\textit{4) Analysis of Context-Probe Usage:}
CoP-FSCIL uses context probes throughout its pipeline. To enlarge intra-class context dispersion, we select reference context statistics according to the farthest context-statistic distance. As shown in Table~\ref{tab:pseudo_context_transfer}, we compare different context-reference selection strategies during base-session training on HapTex. The results show that constructing context probes with the farthest context statistics brings larger performance gains than using randomly selected statistics. This indicates that explicitly simulating excessive intra-class context dispersion in the training data is effective for improving context robustness.

To further analyze the role of context probes and the subsequent modules, we apply Context-Probing Intervention with different coverage levels to support and query samples in the pseudo-incremental stage following the baseline protocol on HapTex, as shown in Table~\ref{tab:cpi_coverage}. Here, $100\%$ denotes applying CPI to all samples, while $60\%$ denotes applying CPI to a randomly selected $60\%$ of the samples. The results show that applying $60\%$-CPI to support samples and $100\%$-CPI to query samples achieves the best performance, whereas the reverse configuration leads to the worst performance. In general, $60\%$-CPI on support samples consistently outperforms $100\%$-CPI on support samples, while $100\%$-CPI on query samples performs better than $60\%$-CPI on query samples. This observation suggests that, under the few-shot support setting, moderately simulating intra-class context dispersion is more appropriate. Excessive perturbation of scarce support samples may instead impair representation learning. In contrast, for query samples, applying stronger context perturbations can better expose excessive intra-class context dispersion, thereby providing stronger generalization and robustness cues for PCQA and PSPC within the CoP-FSCIL framework.

\begin{figure}[!t]
  \centering
    \centerline{\includegraphics[width=\columnwidth]{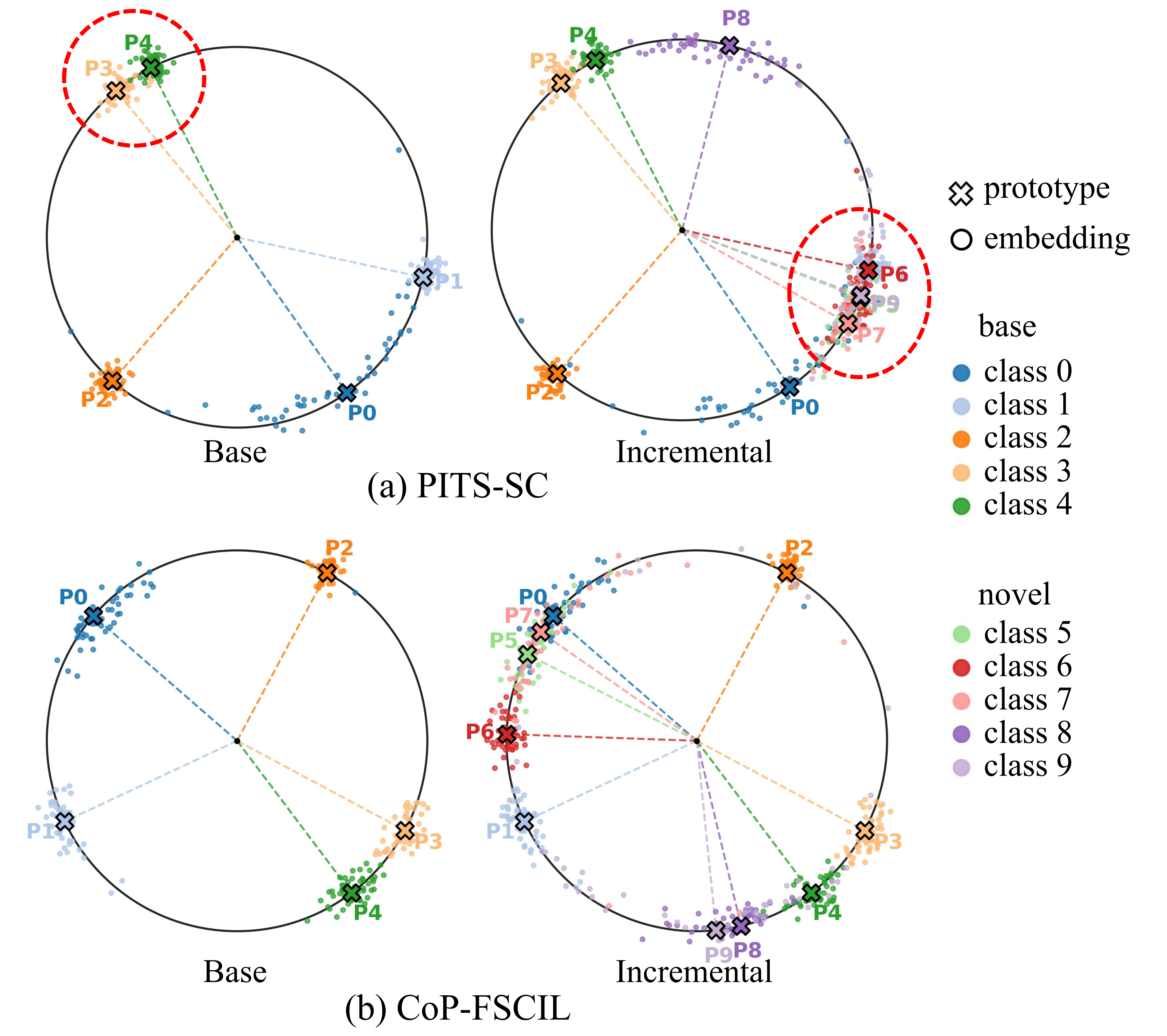}}
    \caption{Embedding-space visualization of PITS-SC and CoP-FSCIL in a two-dimensional angular subspace. Five base classes and five novel classes from HapTex are projected by PCA and normalized onto the unit circle. Compared with PITS-SC, CoP-FSCIL forms more compact intra-class distributions and reduces prototype entanglement in the incremental session.
}
    \label{fig:TEST}
\end{figure}

\textit{5) Embedding Space Visualization:}
We further analyze the adaptability of the embedding space by visualizing the feature distributions before and after applying CoP-FSCIL. Since the cosine classifier is more sensitive to the angular direction of embeddings, we project the embeddings into a two-dimensional angular subspace using PCA and normalize them onto the unit circle for visualization. For a representative visualization, we randomly select five base classes and five novel classes from one incremental session on HapTex, as shown in \cref{fig:TEST}. In the base session, CoP-FSCIL produces more compact intra-class embeddings, as observed for class 0. It also enlarges the angular separation between prototypes of different classes, such as classes 3 and 4. In the incremental session, the PITS-SC baseline is evidently affected by acquisition context. The prototypes of classes 1, 5, 6, 7, and 9 become entangled within a narrow angular region and even overlap in some cases. In contrast, CoP-FSCIL alleviates such prototype entanglement. Although several class directions remain close, their angular separation becomes sufficient for discrimination.

This improvement can be attributed to two factors. First, CoP-FSCIL corrects the relative directions of embeddings for certain classes, such as class 9, by reducing their overfitting to acquisition context and forming a more reasonable candidate region for prototype estimation. Second, CoP-FSCIL calibrates prototypes by weighting support samples according to their probe stability. Nevertheless, these mechanisms may inevitably down-weight or discard highly context-sensitive samples. As a result, a few embeddings deviate substantially from the prototype direction of their corresponding classes.

\begin{figure}[!t]
  \centering
    \centerline{\includegraphics[width=\columnwidth]{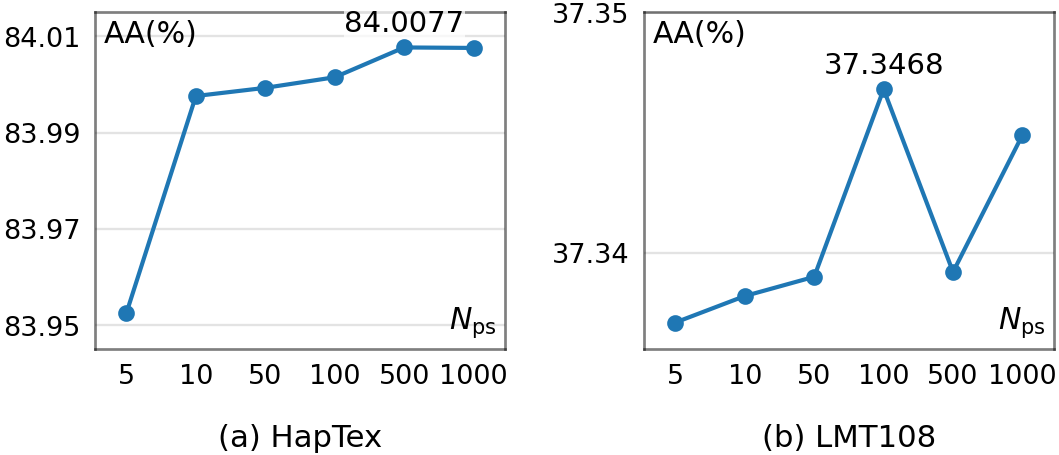}}
    \caption{Sensitivity analysis of the PSPC hyperparameter \(N_{\mathrm{ps}}\) on CoP-FSCIL performance.
}
    \label{caotu6}
\end{figure}

\textit{6) Hyperparameter Sensitivity Analysis:}
We study how the number of generated context probes \(N_{\mathrm{ps}}\), which is used to estimate the probe-stability uncertainty in PSPC, affects performance. To further analyze its influence on both performance and efficiency, we conduct experiments with different values of \(N_{\mathrm{ps}}\). The results are reported in \cref{caotu6} and Table~\ref{tab:nps_update_time}, where Incremental Update Time includes both prototype update time and incremental fine-tuning time. The performance slightly improves as \(N_{\mathrm{ps}}\) increases and becomes relatively stable after reaching a sufficient value. However, further increasing \(N_{\mathrm{ps}}\) brings only marginal gains while introducing additional computational cost. In particular, when  \(N_{\mathrm{ps}}>100\), the Incremental Update Time becomes prohibitively high. Therefore, considering the sample size of each dataset, we set \(N_{\mathrm{ps}}\) to 100 for HapTex and 10 for LMT108 as a practical trade-off between performance and efficiency.

\begin{table}[!t]
\centering
\caption{Incremental update time under different values of $N_{\mathrm{ps}}$.}
\label{tab:nps_update_time}
\renewcommand{\arraystretch}{1.10}
\setlength{\tabcolsep}{2.8pt}
\footnotesize
\begin{tabular*}{\columnwidth}{@{\extracolsep{\fill}}lcccccc@{}}
\toprule
\multirow{2}{*}{Dataset} 
& \multicolumn{6}{c}{Incremental Update Time (s) / $N_{\mathrm{ps}}$} \\
\cmidrule(lr){2-7}
& 5 & 10 & 50 & 100 & 500 & 1000 \\
\midrule
HapTex & 0.10 & 0.17 & 0.41 & 0.86 & 3.95 & 6.52 \\
LMT108 & 0.50 & 0.74 & 2.69 & 5.31 & 24.42 & 47.09 \\
\bottomrule
\end{tabular*}
\end{table}
\section{Conclusion}
\label{Conclusion}

This paper investigates FSCIL under acquisition-context variations,  a prevalent yet challenging problem in emerging sensory modalities such as tactile sensing. We propose CoP-FSCIL, a novel context-probing framework centered on the construction of context probes. Specifically, CoP-FSCIL employs PCQA to perform sample-adaptive probe-response analysis and extract probe-invariant representations, while PSPC further calibrates the classifier by compensating for residual representation uncertainty. Overall, this design enables robust representation and classifier learning under limited context coverage. Experiments on tactile FSCIL benchmarks show consistent improvements over representative baselines, and extended audio results further indicate the generality of context probing. Despite these gains, CoP-FSCIL still relies on multiple probes for reliability estimation. Future work will explore adaptive probe selection and physics-aware probe generation for more efficient and deployable tactile FSCIL. 

\bibliographystyle{IEEEtran}

\bibliography{IEEEexample}

\begin{thebibliography}{10}
\providecommand{\url}[1]{#1}
\csname url@samestyle\endcsname
\providecommand{\newblock}{\relax}
\providecommand{\bibinfo}[2]{#2}
\providecommand{\BIBentrySTDinterwordspacing}{\spaceskip=0pt\relax}
\providecommand{\BIBentryALTinterwordstretchfactor}{4}
\providecommand{\BIBentryALTinterwordspacing}{\spaceskip=\fontdimen2\font plus
\BIBentryALTinterwordstretchfactor\fontdimen3\font minus \fontdimen4\font\relax}
\providecommand{\BIBforeignlanguage}[2]{{%
\expandafter\ifx\csname l@#1\endcsname\relax
\typeout{** WARNING: IEEEtran.bst: No hyphenation pattern has been}%
\typeout{** loaded for the language `#1'. Using the pattern for}%
\typeout{** the default language instead.}%
\else
\language=\csname l@#1\endcsname
\fi
#2}}
\providecommand{\BIBdecl}{\relax}
\BIBdecl

\bibitem{li2025anchorinv}
C.~Li, B.~Gao, G.~D. Jones, T.~Denison, and T.~Zhu, ``Anchorinv: Few-shot class-incremental learning of physiological signals via feature space-guided inversion,'' in \emph{Proc. AAAI Conf. Artif. Intell.}, vol.~39, no.~13, 2025, pp. 14\,274--14\,282.

\bibitem{cao2025few}
L.~Cao, H.~Li, Y.~Dong, T.~Liu, and J.~Li, ``Few-shot class-incremental learning with dynamic prototype refinement for brain activity classification,'' \emph{IEEE J. Biomed. Health Inform.}, 2025.

\bibitem{li2025fewaudio}
Y.~Li, W.~Cao, J.~Tan, Q.~Li, and G.~Chen, ``Few-shot class-incremental audio classification using pseudoincrementally trained embedding learner and continually updated stochastic classifier,'' \emph{IEEE/ACM Trans. Audio, Speech, Lang. Process.}, 2025.

\bibitem{li2024efficient}
D.~Li, A.~Zhang, J.~Gao, and B.~Qi, ``An efficient memory module for graph few-shot class-incremental learning,'' \emph{Adv. Neural Inf. Process. Syst.}, vol.~37, pp. 130\,084--130\,108, 2024.

\bibitem{wei2025few}
L.~Wei, Y.~Ma, Z.~Lin, F.~Wang, C.~Jin, H.~Zhao, and D.~Chen, ``Few-shot incremental multi-modal learning via touch guidance and imaginary vision synthesis,'' in \emph{Proc. Int. Joint Conf. Artif. Intell.}, 2025, pp. 2045--2053.

\bibitem{xiang2025seeing}
T.~Xiang, X.~Xu, B.~Liu, J.~Li, Y.~Li, and S.~He, ``Seeing 3d through 2d lenses: 3d few-shot class-incremental learning via cross-modal geometric rectification,'' in \emph{Proc. IEEE/CVF Int. Conf. Comput. Vis.}, 2025, pp. 6761--6771.

\bibitem{peng2022few}
C.~Peng, K.~Zhao, T.~Wang, M.~Li, and B.~C. Lovell, ``Few-shot class-incremental learning from an open-set perspective,'' in \emph{Proc. Eur. Conf. Comput. Vis.}\hskip 1em plus 0.5em minus 0.4em\relax Springer, 2022, pp. 382--397.

\bibitem{guo2023decision}
C.~Guo, Q.~Zhao, S.~Lyu, B.~Liu, C.~Wang, L.~Chen, and G.~Cheng, ``Decision boundary optimization for few-shot class-incremental learning,'' in \emph{Proc. IEEE/CVF Int. Conf. Comput. Vis.}, 2023, pp. 3501--3511.

\bibitem{li2025adaptive}
L.~Li, Y.~Tan, S.~Yang, H.~Cheng, Y.~Dong, and L.~Yang, ``Adaptive decision boundary for few-shot class-incremental learning,'' in \emph{Proc. AAAI Conf. Artif. Intell.}, vol.~39, no.~17, 2025, pp. 18\,359--18\,367.

\bibitem{agarwal2022semantics}
A.~Agarwal, B.~Banerjee, F.~Cuzzolin, and S.~Chaudhuri, ``Semantics-driven generative replay for few-shot class incremental learning,'' in \emph{Proc. ACM Int. Conf. Multimedia}, 2022, pp. 5246--5254.

\bibitem{song2023learning}
Z.~Song, Y.~Zhao, Y.~Shi, P.~Peng, L.~Yuan, and Y.~Tian, ``Learning with fantasy: Semantic-aware virtual contrastive constraint for few-shot class-incremental learning,'' in \emph{Proc. IEEE/CVF Conf. Comput. Vis. Pattern Recognit.}, 2023, pp. 24\,183--24\,192.

\bibitem{ahmed2024orco}
N.~Ahmed, A.~Kukleva, and B.~Schiele, ``Orco: Towards better generalization via orthogonality and contrast for few-shot class-incremental learning,'' in \emph{Proc. IEEE/CVF Conf. Comput. Vis. Pattern Recognit.}, 2024, pp. 28\,762--28\,771.

\bibitem{kim2025can}
J.~Kim, Y.~Ku, and S.~Baek, ``Can synthetic images conquer forgetting? beyond unexplored doubts in few-shot class-incremental learning,'' in \emph{Proc. IEEE/CVF Int. Conf. Comput. Vis.}, 2025, pp. 5214--5223.

\bibitem{park2024pre}
K.-H. Park, K.~Song, and G.-M. Park, ``Pre-trained vision and language transformers are few-shot incremental learners,'' in \emph{Proc. IEEE/CVF Conf. Comput. Vis. Pattern Recognit.}, 2024, pp. 23\,881--23\,890.

\bibitem{liu2025sec}
Y.~Liu and M.~Yang, ``Sec-prompt: Semantic complementary prompting for few-shot class-incremental learning,'' in \emph{Proc. IEEE/CVF Conf. Comput. Vis. Pattern Recognit.}, 2025, pp. 25\,643--25\,656.

\bibitem{lrt}
Y.~Zhao, J.~Li, Z.~Song, and Y.~Tian, ``Language-inspired relation transfer for few-shot class-incremental learning,'' \emph{IEEE Trans. Pattern Anal. Mach. Intell.}, vol.~47, no.~2, pp. 1089--1102, 2025.

\bibitem{tao2020few}
X.~Tao, X.~Hong, X.~Chang, S.~Dong, X.~Wei, and Y.~Gong, ``Few-shot class-incremental learning,'' in \emph{Proc. IEEE/CVF Conf. Comput. Vis. Pattern Recognit.}, 2020, pp. 12\,183--12\,192.

\bibitem{zhang2021few}
C.~Zhang, N.~Song, G.~Lin, Y.~Zheng, P.~Pan, and Y.~Xu, ``Few-shot incremental learning with continually evolved classifiers,'' in \emph{Proc. IEEE/CVF Conf. Comput. Vis. Pattern Recognit.}, 2021, pp. 12\,455--12\,464.

\bibitem{zhou2022forward}
D.-W. Zhou, F.-Y. Wang, H.-J. Ye, L.~Ma, S.~Pu, and D.-C. Zhan, ``Forward compatible few-shot class-incremental learning,'' in \emph{Proc. IEEE/CVF Conf. Comput. Vis. Pattern Recognit.}, 2022, pp. 9046--9056.

\bibitem{zou2024compositional}
Y.~Zou, S.~Zhang, H.~Zhou, Y.~Li, and R.~Li, ``Compositional few-shot class-incremental learning,'' in \emph{Proc. Int. Conf. Mach. Learn.}\hskip 1em plus 0.5em minus 0.4em\relax PMLR, 2024, pp. 62\,964--62\,977.

\bibitem{shi2025lark}
J.~Shi, J.~Zhao, Y.~Yang, X.~Wu, J.~Li, and L.~He, ``Lark: Low-rank updates after knowledge localization for few-shot class-incremental learning,'' in \emph{Proc. IEEE/CVF Int. Conf. Comput. Vis.}, 2025, pp. 3607--3617.

\bibitem{xie2025flexi}
W.~Xie, Y.~Wang, C.~Liu, Z.~Jiang, and X.~Yang, ``Flexi-fscil: Adaptive knowledge retention for breaking the stability-plasticity dilemma in few-shot class-incremental learning,'' in \emph{Proc. IEEE/CVF Int. Conf. Comput. Vis.}, 2025, pp. 2451--2460.

\bibitem{krizhevsky2009learning}
A.~Krizhevsky, G.~Hinton \emph{et~al.}, ``Learning multiple layers of features from tiny images,'' 2009.

\bibitem{wah2011caltech}
C.~Wah, S.~Branson, P.~Welinder, P.~Perona, and S.~Belongie, ``The caltech-ucsd birds-200-2011 dataset,'' 2011.

\bibitem{russakovsky2015imagenet}
O.~Russakovsky, J.~Deng, H.~Su, J.~Krause, S.~Satheesh, S.~Ma, Z.~Huang, A.~Karpathy, A.~Khosla, M.~Bernstein \emph{et~al.}, ``Imagenet large scale visual recognition challenge,'' \emph{Int. J. Comput. Vis.}, vol. 115, no.~3, pp. 211--252, 2015.

\bibitem{ma2022few}
T.-F. Ma, W.-L. Zheng, and B.-L. Lu, ``Few-shot class-incremental learning for eeg-based emotion recognition,'' in \emph{Proc. Int. Conf. Neural Inf. Process.}\hskip 1em plus 0.5em minus 0.4em\relax Springer, 2022, pp. 445--455.

\bibitem{sun2023few}
L.~Sun, M.~Zhang, B.~Wang, and P.~Tiwari, ``Few-shot class-incremental learning for medical time series classification,'' \emph{IEEE J. Biomed. Health Inform.}, vol.~28, no.~4, pp. 1872--1882, 2023.

\bibitem{gao2026tape}
Y.~Gao, W.~Liang, G.~Wang, S.~Guan, L.~Zong, D.~Zhang, and X.~Liu, ``Tape: Task-adaptive prototype evolution in audio-language models for fully few-shot class-incremental audio classification,'' in \emph{Proc. IEEE/CVF Conf. Comput. Vis. Pattern Recognit.}, 2026, pp. 19\,570--19\,579.

\bibitem{10.1145/3754452}
H.~Fu, F.~Yang, B.~Wang, W.~Ji, H.~Zhao, C.~Zhang, R.~Zimmermann, and H.~Qian, ``Visuo-tactile class-incremental learning,'' \emph{ACM Trans. Multimedia Comput. Commun. Appl.}, vol.~21, no.~11, Nov. 2025.

\bibitem{visgel}
Y.~Li, J.-Y. Zhu, R.~Tedrake, and A.~Torralba, ``Connecting touch and vision via cross-modal prediction,'' in \emph{Proc. IEEE/CVF Conf. Comput. Vis. Pattern Recognit.}, 2019, pp. 10\,609--10\,618.

\bibitem{touchandgo}
F.~Yang, C.~Ma, J.~Zhang, J.~Zhu, W.~Yuan, and A.~Owens, ``Touch and go: learning from human-collected vision and touch,'' in \emph{Adv. Neural Inf. Process. Syst.}, 2022, pp. 8081--8103.

\bibitem{anytouch}
R.~Feng, J.~Hu, W.~Xia, A.~Shen, Y.~Sun, B.~Fang, D.~Hu \emph{et~al.}, ``Anytouch: Learning unified static-dynamic representation across multiple visuo-tactile sensors,'' in \emph{Proc. Int. Conf. Learn. Represent.}, 2025.

\bibitem{tvl}
L.~Fu, G.~Datta, H.~Huang, W.~C.-H. Panitch, J.~Drake, J.~Ortiz, M.~Mukadam, M.~Lambeta, R.~Calandra, and K.~Goldberg, ``A touch, vision, and language dataset for multimodal alignment,'' in \emph{Proc. Int. Conf. Mach. Learn.}\hskip 1em plus 0.5em minus 0.4em\relax PMLR, 2024, pp. 14\,080--14\,101.

\bibitem{touch100k}
N.~Cheng, J.~Xu, C.~Guan, J.~Gao, W.~Wang, Y.~Li, F.~Meng, J.~Zhou, B.~Fang, and W.~Han, ``Touch100k: A large-scale touch-language-vision dataset for touch-centric multimodal representation,'' \emph{Inf. Fusion}, vol. 124, p. 103305, 2025.

\bibitem{lmt108}
M.~Strese, Y.~Boeck, and E.~Steinbach, ``Content-based surface material retrieval,'' in \emph{Proc. IEEE World Haptics Conf.}\hskip 1em plus 0.5em minus 0.4em\relax IEEE, 2017, pp. 352--357.

\bibitem{haptex}
J.~Jiao, Y.~Zhang, D.~Wang, X.~Guo, and X.~Sun, ``Haptex: A database of fabric textures for surface tactile display,'' in \emph{Proc. IEEE World Haptics Conf.}\hskip 1em plus 0.5em minus 0.4em\relax IEEE, 2019, pp. 331--336.

\bibitem{eguchi2026cluster}
M.~Eguchi, T.~Hayase, Y.~Hiroi, and T.~Hiraki, ``Cluster haptic texture dataset: Haptic texture dataset with varied velocity-direction sliding contacts,'' \emph{Sci. Data}, 2026.

\bibitem{8605315}
O.~Holland, E.~Steinbach, R.~V. Prasad, Q.~Liu, Z.~Dawy, A.~Aijaz, N.~Pappas, K.~Chandra, V.~S. Rao, S.~Oteafy, M.~Eid, M.~Luden, A.~Bhardwaj, X.~Liu, J.~Sachs, and J.~Araújo, ``The ieee 1918.1 “tactile internet” standards working group and its standards,'' \emph{Proc. IEEE}, vol. 107, no.~2, pp. 256--279, 2019.

\bibitem{liu2017recent}
H.~Liu, Y.~Wu, F.~Sun, and D.~Guo, ``Recent progress on tactile object recognition,'' \emph{Int. J. Adv. Robot. Syst.}, vol.~14, no.~4, p. 1729881417717056, 2017.

\bibitem{huang2021texture}
S.~Huang and H.~Wu, ``Texture recognition based on perception data from a bionic tactile sensor,'' \emph{Sensors}, vol.~21, no.~15, p. 5224, 2021.

\bibitem{gandarias2019cnn}
J.~M. Gandarias, A.~J. Garcia-Cerezo, and J.~M. Gomez-de Gabriel, ``Cnn-based methods for object recognition with high-resolution tactile sensors,'' \emph{IEEE Sensors J.}, vol.~19, no.~16, pp. 6872--6882, 2019.

\bibitem{donato2025tactile}
E.~Donato, D.~Pelliccia, M.~Hosseinzadeh, M.~Amiri, and E.~Falotico, ``Tactile object recognition with recurrent neural networks through a perceptive soft gripper,'' \emph{IEEE Robot. Autom. Lett.}, 2025.

\bibitem{xie2024deep}
Y.~Xie, H.~Cheng, C.~Yuan, L.~Zheng, Z.~Peng, and B.~Meng, ``Deep learning-assisted object recognition with hybrid triboelectric-capacitive tactile sensor,'' \emph{Microsyst. Nanoeng.}, vol.~10, no.~1, p. 165, 2024.

\bibitem{zhao2024augmented}
X.~Zhao, Z.~Sun, and C.~Lee, ``Augmented tactile perception of robotic fingers enabled by ai-enhanced triboelectric multimodal sensors,'' \emph{Adv. Funct. Mater.}, vol.~34, no.~49, p. 2409558, 2024.

\bibitem{cao2024multimodal}
G.~Cao, J.~Jiang, D.~Bollegala, M.~Li, and S.~Luo, ``Multimodal zero-shot learning for tactile texture recognition,'' \emph{Robot. Auton. Syst.}, vol. 176, p. 104688, 2024.

\bibitem{ueda2024visuo}
S.~Ueda, A.~Hashimoto, M.~Hamaya, K.~Tanaka, and H.~Saito, ``Visuo-tactile zero-shot object recognition with vision-language model,'' in \emph{Proc. IEEE/RSJ Int. Conf. Intell. Robots Syst.}\hskip 1em plus 0.5em minus 0.4em\relax IEEE, 2024, pp. 7243--7250.

\bibitem{arp}
W.~Xie, Y.~Li, Q.~He, W.~Cao, and T.~Virtanen, ``Few-shot class-incremental audio classification using adaptively-refined prototypes,'' in \emph{Proc. Interspeech}.\hskip 1em plus 0.5em minus 0.4em\relax International Speech Communication Association, 2023, pp. 301--305.

\bibitem{ls}
V.~Panayotov, G.~Chen, D.~Povey, and S.~Khudanpur, ``Librispeech: an asr corpus based on public domain audio books,'' in \emph{Proc. IEEE Int. Conf. Acoust., Speech Signal Process.}\hskip 1em plus 0.5em minus 0.4em\relax IEEE, 2015, pp. 5206--5210.

\bibitem{ns}
J.~Engel, C.~Resnick, A.~Roberts, S.~Dieleman, M.~Norouzi, D.~Eck, and K.~Simonyan, ``Neural audio synthesis of musical notes with wavenet autoencoders,'' in \emph{Proc. Int. Conf. Mach. Learn.}\hskip 1em plus 0.5em minus 0.4em\relax PMLR, 2017, pp. 1068--1077.

\bibitem{li_2026_INTERSPEECH}
Y.~Li, G.~Chen, Q.~Li, and S.~Huang, ``Few-shot class-variable incremental audio classification via prototype adaptation and pseudo class-variable training,'' in \emph{Proc. Interspeech}.\hskip 1em plus 0.5em minus 0.4em\relax International Speech Communication Association, 2026.

\bibitem{ldc}
B.~Liu, B.~Yang, L.~Xie, R.~Wang, Q.~Tian, and Q.~Ye, ``Learnable distribution calibration for few-shot class-incremental learning,'' \emph{IEEE Trans. Pattern Anal. Mach. Intell.}, vol.~45, no.~10, pp. 12\,699--12\,706, 2023.

\bibitem{dsn}
B.~Yang, M.~Lin, Y.~Zhang, B.~Liu, X.~Liang, R.~Ji, and Q.~Ye, ``Dynamic support network for few-shot class incremental learning,'' \emph{IEEE Trans. Pattern Anal. Mach. Intell.}, vol.~45, no.~3, pp. 2945--2951, 2023.

\bibitem{pan}
Y.~Li, W.~Cao, W.~Xie, J.~Li, and E.~Benetos, ``Few-shot class-incremental audio classification using dynamically expanded classifier with self-attention modified prototypes,'' \emph{IEEE Trans. Multimedia}, vol.~26, pp. 1346--1360, 2024.

\bibitem{amfo}
Y.~Li, J.~Li, Y.~Si, J.~Tan, and Q.~He, ``Few-shot class-incremental audio classification with adaptive mitigation of forgetting and overfitting,'' \emph{IEEE/ACM Trans. Audio, Speech, Lang. Process.}, vol.~32, pp. 2297--2311, 2024.

\bibitem{pengi}
S.~Deshmukh, B.~Elizalde, R.~Singh, and H.~Wang, ``{Pengi}: an audio language model for audio tasks,'' in \emph{Adv. Neural Inf. Process. Syst.}, vol.~36.\hskip 1em plus 0.5em minus 0.4em\relax Curran Associates, Inc., 2023, pp. 18\,090--18\,108.

\end{thebibliography}

\end{document}